\documentclass[final,twocolumn]{cvpr}

\usepackage{times}
\usepackage{epsfig}
\usepackage{graphicx}
\usepackage{amsmath}
\usepackage{amssymb}
\usepackage{subcaption}
\usepackage{multirow}
\usepackage{appendix}

\usepackage[pagebackref=true,breaklinks=true,colorlinks,bookmarks=false]{hyperref}

\def\cvprPaperID{8502} 
\def\confYear{CVPR 2021}
\begin{document}


\title{Contrastive Learning for Unsupervised Image-to-Image Translation}

\author{
    Hanbit Lee \qquad Jinseok Seol \qquad Sang-goo Lee \\
    Department of Computer Science and Engineering \\
    Seoul National University \\
    {\tt\small \{skcheon, jamie, sglee\}@europa.snu.ac.kr}
}


\twocolumn[{%
    \renewcommand\twocolumn[1][]{#1}%
    \maketitle
    \begin{center}
        \includegraphics[width=.95\linewidth]{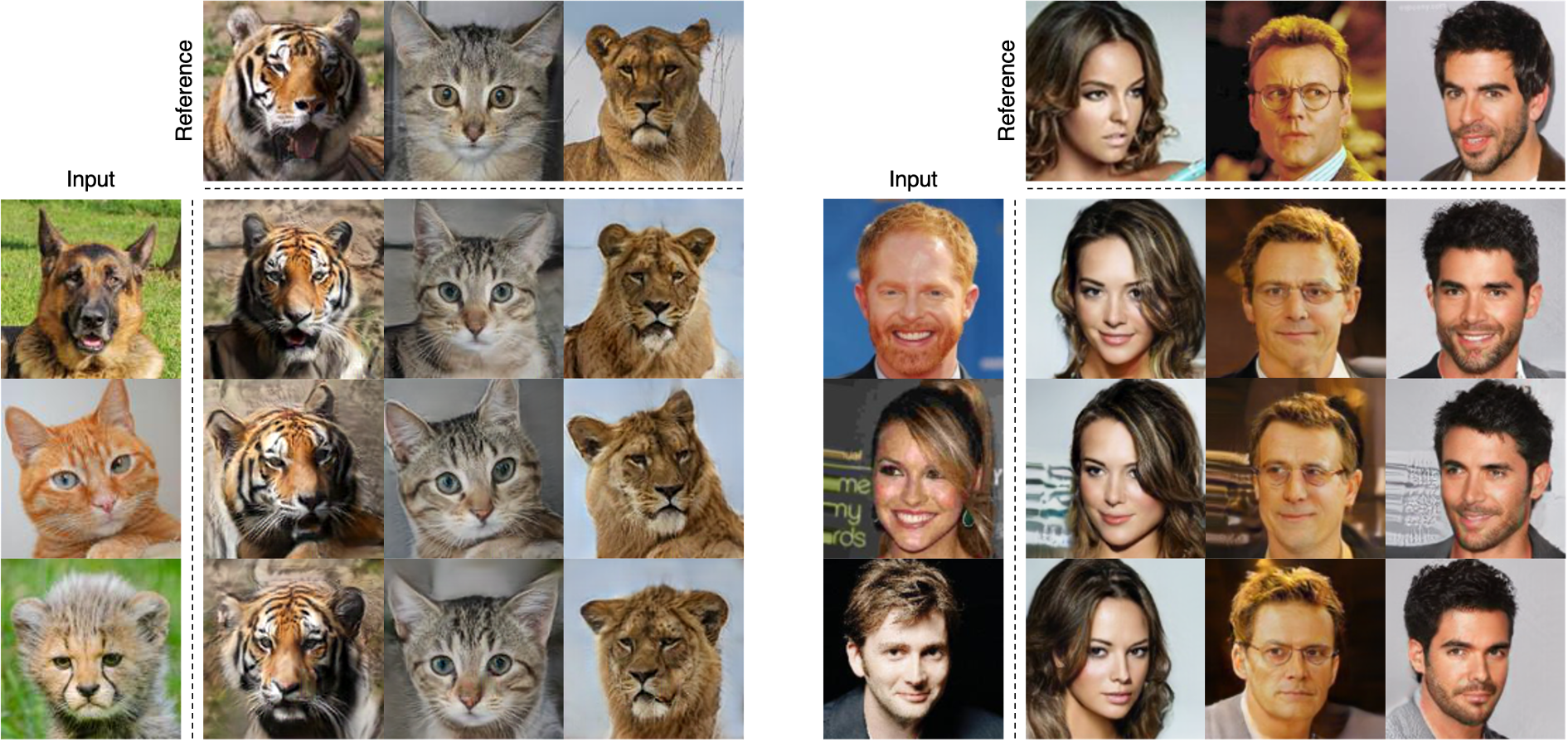}
        \captionof{figure}{Unsupervised image-to-image translation results on the AFHQ dataset (left) and CelebA dataset (right). First column and first row shows the input images and the reference images, respectively, while the rest of the images are synthesized by CLUIT. CLUIT can create high-quality images reflecting the style of given arbitrary reference image. Note that we did not use any ground truth labels.}
        \vspace{+0.5em}
        \label{fig:teaser}
    \end{center}
}]

\begin{abstract}
    Image-to-image translation aims to learn a mapping between different groups of visually distinguishable images.
    While recent methods have shown impressive ability to change even intricate appearance of images, they still rely on domain labels in training a model to distinguish between distinct visual features.
    Such dependency on labels often significantly limits the scope of applications since consistent and high-quality labels are expensive.
    Instead, we wish to capture visual features from images themselves and apply them to enable realistic translation without human-generated labels.
    To this end, we propose an unsupervised image-to-image translation method based on contrastive learning.
    The key idea is to learn a discriminator that differentiates between distinctive styles and let the discriminator supervise a generator to transfer those styles across images.
    During training, we randomly sample a pair of images and train the generator to change the appearance of one towards another while keeping the original structure.
    Experimental results show that our method outperforms the leading unsupervised baselines in terms of visual quality and translation accuracy.
\end{abstract}


\section{Introduction}

    Image-to-image translation aims to learn a mapping between different groups of visually distinguishable images.
    Based on the success of generative adversarial networks \cite{goodfellow2014generative}, recent studies have shown a remarkable ability to learn and render intricate appearance changes across image classes \cite{zhu2017unpaired, kim2017learning, yi2017dualgan, liu2017unsupervised, choi2018stargan, liu2019few, choi2020stargan}.
    These approaches rely on domain labels to specify visually different groups, which often significantly limits the scope of applications.
    It's because inconsistent or coarse labels easily degrade translation results while collecting high-quality labels is usually expensive.

    To overcome such limitations, more recent studies explore an unsupervised approach \cite{bahng2020exploring, baek2020rethinking}.
    These methods leverage image clustering techniques to identify distinct groups of images without ground-truth labels.
    Specifically, they produce pseudo-labels of unlabeled images by clustering algorithms and learn a mapping between these estimated groups using a multi-domain discriminator.
    While these methods have shown promising results, it is usually difficult to produce accurate clusters when the data distribution is unbalanced or the predefined configurations are misleading.
    For example, the clustering methods require the user to specify the number of clusters, which is generally difficult to choose without prior knowledge, despite having a significant impact on the results.

    Instead, we propose an alternative, rather straightforward way to circumvent these limitations.
    Our key idea is to supervise the image translation process with visual similarity captured by contrastive learning \cite{hadsell2006dimensionality, he2020momentum, chen2020simple} instead of learning separate domain distribution.
    Contrastive learning aims to learn a useful representation by contrasting between positive pairs and negative pairs.
    Through the contrasting process, the representations of similar samples are mapped closely together while dissimilar samples are mapped further away.
    We observe that the contrastive learning captures meaningful visual features that form the unique appearance of individual image, which we call a \textit{style}.
    Conversely, we call the characteristics shared among all images (e.g., pose) a \textit{structure}.

    We enable effective, unsupervised image-to-image translation using contrastive learning to transfer styles between images.
    Specifically, we introduce a contrastive discriminator that has two branches, one for contrastive representation and another for typical GAN logit.
    During training, the discriminator is provided with not only real and fake images but also an augmented version of real images to learn the contrastive representation.
    The representation is utilized to induce an output image to have a style similar to a given reference image.
    Furthermore, we introduce a patch aggregation method that effectively mitigates the structure loss problem of output images.

    Empirical results demonstrate the advantage of our method over existing unsupervised baselines in terms of visual quality and translation accuracy.
    Moreover, we analyze the latent style space to verify whether the space learns semantically meaningful styles.
    Through extensive ablation studies, we investigate the effect of individual components including data augmentation policy and contrastive style matching strategies for further insights.


\section{Related Work}

    \subsection{Multi-domain image-to-image translation}

        Most of the existing image-to-image translation methods are based on the conditional GAN \cite{mirza2014conditional}.
        They learn domain-specific discriminators to identify whether the generated image belongs to the target domain while imposing additional constraints like cycle consistency \cite{zhu2017unpaired} to preserve domain-invariant characteristics.
        While earlier methods consider a bijective mapping between two domains \cite{zhu2017unpaired, liu2017unsupervised, kim2017learning, yi2017dualgan}, more recent works present multi-domain methods that enable translation across multiple domains with a unified model \cite{choi2018stargan, liu2019few, choi2020stargan}.
        These methods employ an auxiliary domain classifier \cite{choi2018stargan} or a multi-task discriminator \cite{liu2019few, choi2020stargan} for scalable multi-domain translation.

        While successful, these methods rely on a vast quantity of domain labels, which often becomes a serious bottleneck.
        To reduce such appetite for labels, more recent studies propose fully unsupervised methods that leverage pseudo-labels acquired by the image clustering methods \cite{bahng2020exploring, baek2020rethinking}.
        However these methods easily yield unintended translation results if the clustering algorithms fail to produce consistent clusters.
        Our work differs in that we present a model that implicitly learns and transfers visual features without explicit domain separation.

    \subsection{Style transfer}

        Our work can be seen as a photorealistic style transfer task as we aim to transfer the style of an image to another while preserving its structure.
        Photorealistic style transfer methods attempt to transform an image into its stylized version reflecting arbitrary reference images \cite{luan2017deep, li2018closed, yoo2019photorealistic}.
        However, these works presume style as color distributions and mainly focus on local color change rather than capturing semantically meaningful features.
        Recently, SwappingAutoencoder \cite{park2020swapping} presented a code swap-based autoencoder that transfers texture information between images by learning co-occurrence patch statistics between images.
        While it succeeds in capturing local semantic features, it often fails to capture higher-level semantics such as the complex appearance of animals, since the model highly relies on the local patch statistics.
        In contrast, our model learns distinctive visual features that form higher-level semantics by contrastive learning.

\begin{figure*}[!htbp]
    \centering
    \includegraphics[width=1.0\linewidth]{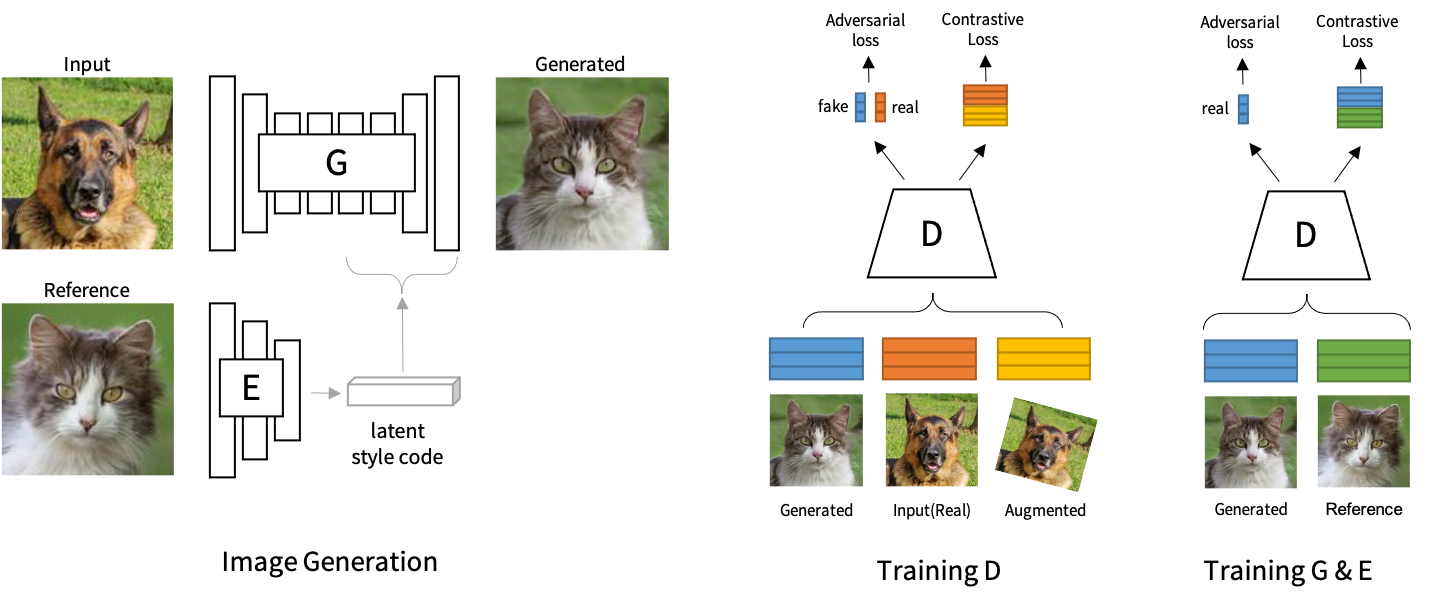}
    \caption{An overview of our proposed method, CLUIT. The method consists of three modules: a generator $G$, a style encoder $E$, and a contrastive discriminator $D$. Using a latent style code from the encoder, the generator synthesizes an output image. Input and reference images are randomly sampled within a batch when training. The discriminator learns the real image distribution from adversarial loss. At the same time, it exploits data augmentation techniques to learn image representations through contrastive loss. The generator is guided to synthesize an image with a visual style similar to the reference image through contrastive match loss.}
    \label{fig:overview}
\end{figure*}

    \subsection{Unsupervised representation learning}

        Our work is also closely related to unsupervised representation learning techniques in that we aim to capture and transfer such meaningful visual features without supervision.
        Various pretext tasks are designed for this purpose, such as image reconstruction \cite{hinton2006reducing}, denoising \cite{vincent2008extracting}, inpainting \cite{pathak2016context}, colorization \cite{larsson2017colorization}, context prediction \cite{noroozi2016unsupervised, santa2017deeppermnet}, and geometric prediction \cite{noroozi2017representation, liu2019exploiting}.
        More recent approaches focus on learning representations that maximize the mutual information between examples that are assumed to be closely related.
        These methods employ a contrastive loss to attract the associated examples closer than other negatives \cite{oord2018representation, bachman2019learning, he2020momentum, chen2020simple}.
        Through linear evaluation and various down-stream tasks, contrastive learning has shown the ability to capture complex visual features.
        In this work, we present a method that effectively utilizes those capabilities in an image-to-image translation setting, which has not been explored before.


\section{Method}

    Given an unlabeled set of images $X \subset \mathbb{R}^{H \times W \times C}$, we aim to translate an input image $x_o \in X$ to appear like a reference image $x_r \in X$ while preserving the class-invariant characteristics (i.e., pose).
    To this end, we propose a contrastive learning based unsupervised image-to-image translation framework, namely \textit{CLUIT}, consisting of three modules: (1) a contrastive discriminator $D$ that learns visual styles of images while guiding a generator to synthesize realistic images, (2) a style encoder $E$ that encodes an image into latent style code and (3) a generator $G$ that translates the input image $x_o$ into a corresponding output reflecting the encoded style of the reference image $x_r$.
    Figure \ref{fig:overview} illustrates an overview of our framework.

    \subsection{Contrastive discriminator}

        Our discriminator $D$ is designed to have two output branches above shared layers, each of which produces a contrastive representation $v = D_{\text{ct}}(x)$ and a typical GAN logit $D_{\text{adv}}(x)$ for a given image $x$.
        The contrastive representation learns the visual styles of images by maximizing the mutual information between an image $x$ and its augmented version $x^+$ in contrast to other negative images $x^-$ within the dataset.
        Specifically, the images $x$, $x^+$, and $N$ negatives are mapped into $K$-dimensional vectors $v, v^+ \in \mathbb{R}^K$ and $v_i^- \in \mathbb{R}^{K}$, respectively.
        The vectors are normalized to prevent collapsing and then used for ($N + 1$)-way classification as follows:
        \begin{equation} \label{eq:loss-ct-d}
            \mathcal{L}_{\text{ct}}^D = - \log \frac{
                \textstyle \exp(v \cdot v^+ / \tau)
            }{
                \textstyle \exp(v \cdot v^+ / \tau)
                +
                \textstyle \sum_{i=1}^N \exp(v \cdot v^-_i / \tau)
            },
        \end{equation}
        where $\tau$ is a temperature scale factor.
        We maintain a large dictionary of negative examples using a memory bank architecture following MOCO \cite{he2020momentum}.
        The contrastive representation provides proper guidance for the generator to transfer styles between images, which we describe in Section \ref{section:ct}.

        We also impose the adversarial loss to encourage the generator to synthesize realistic images:
        \begin{equation} \label{eq:loss-adv}
            \mathcal{L}_{\text{adv}} =
                \mathbb{E} \left[ \log D_{\text{adv}}(x_o) \right]
                +
                \mathbb{E} \left[\log(1 - D_{\text{adv}}(x_g)) \right],
        \end{equation}
        where $x_g$ is a synthesized image which we describe in the following subsection.

    \subsection{Synthesizing a hybrid image with locality-preserving generator}

        During training, we randomly sample two images from dataset and use one image as an input image and another as a reference image that provides a style reference.
        We notate the input and the reference image as $x_o$ and $x_r$, respectively.
        In practice, we use the first half of batch images as the input images and the rest as the reference images.

        A style encoder $E$ is used to extract the style information from the reference image $x_r$ by encoding the image into corresponding latent style code $t_r = E(x_r)$.
        Our generator $G$ learns to translate the input image $x_o$ into the output image $x_g = G(x_o, t_r)$ using the encoded style code $t_r$.
        We adopt the locality-preserving architecture as our generator, which uses adaptive instance normalization (AdaIN \cite{huang2017arbitrary}) to inject the latent style code $t_r$ into $G$.

        We also generate a reconstructed version of the input image $x_o$ to impose the cycle consistency constraint \cite{zhu2017unpaired} to ensure that the generated image $x_g$ preserves the class-invariant characteristics (i.e., pose) of the input image:
        \begin{equation} \label{eq:loss-cyc}
            \mathcal{L}_{\text{cyc}} = \mathbb{E} \left[ \left\| x_o - G(x_g, t_o) \right\|_1 \right].
        \end{equation}

    \subsection{Style transfer via contrastive style match}
    \label{section:ct}

        Most of the existing methods transfer the class-specific style by learning class-specific discriminators to identify whether the output image is from the given target class.
        We instead impose a style matching constraint defined on the contrastive representation space of the discriminator.
        We use the same form of contrastive loss as used for learning the discriminator but compute the loss using the contrastive representations of the output image $x_g$ and the reference image $x_r$ to guide $x_g$ to have a style similar to $x_r$:
        \begin{equation} \label{eq:loss-ct-g}
            \mathcal{L}_{\text{ct}}^G = \mathbb{E} \left[
                - \log \frac{
                    \textstyle \exp(v_g \cdot v_r / \tau)
                }{
                    \textstyle \exp(v_g \cdot v_r / \tau)
                    +
                    \textstyle \sum_{i=1}^N \exp(v_g \cdot v^-_i / \tau)
                }
            \right],
        \end{equation}
        where $v_g = D_{\text{ct}}(x_g)$ and $v_r = D_{\text{ct}}(x_r)$ denote the respective contrastive representation of $x_g$ and $x_r$.
        The negatives are sampled from the same dictionary used to learn the discriminator.

        Through the contrastive style match, the model can transfer the style of the reference image to the resulting image.
        However, we observe that the model often suffers from the \textit{structure loss} problem where the output images emulate the reference images rather than preserving the structure of the input images (see Figure \ref{fig:matching}).
        To alleviate this problem, we propose using a \textit{patch aggregation} based style match.
        The key idea is to use multiple local patch representations instead of a single full-image representation to prevent the output images from emulating the reference.

        Specifically, $M$ patches of size 1/8 to 1 of the full image dimension are randomly selected from both the output and the reference image, and compute the style match loss using the mean representation of patches in each image:
        \begin{equation}
            v = \frac{1}{M} \sum_{i=1}^M D_{\text{ct}}(x^{(i)}),
        \end{equation}
        where $x^{(i)}$ denote $i$-th patch from image $x$.

    \subsection{Full objective}

        Finally, our full objective functions can be written as:
        \begin{equation} \label{eq:full}
            \begin{split}
                \mathcal{L}_D &= -\mathcal{L}_{\text{adv}} + \lambda_{\text{ct}}^D \mathcal{L}_{\text{ct}}^D,
                \\
                \mathcal{L}_{G, E} &= \mathcal{L}_{\text{adv}} + \lambda_{\text{cyc}} \mathcal{L}_{\text{cyc}} + \lambda_{\text{ct}}^G \mathcal{L}_{\text{ct}}^G,
            \end{split}
        \end{equation}
        where $\lambda_{\text{ct}}^D$, $\lambda_{\text{ct}}^G$ and $\lambda_{\text{cyc}}$ are hyper-parameters for each term.


\section{Experiments}

    In this section, we describe the evaluation setups and experimental results. 
    We compare the proposed method to the state-of-the-art methods and present an additional analysis of our method for further insights.

    \vspace{0.2cm}

    \noindent \textbf{Datasets.}
    We evaluate our method on the following datasets and all the images are resized to $128 \times 128$ resolution for a fair comparison.
    AFHQ \cite{choi2020stargan} contains 15,000 high-quality animal face images which are roughly categorized into three classes (i.e., cat, dog, and wildlife).
    CelebA \cite{liu2015deep} contains 202,599 face images of celebrities annotated with 40 binary facial attributes. We initially center-crop the images to $178 \times 178$ size and then resize them to $128 \times 128$.

\begin{figure*}[!htbp]
    \centering
    \includegraphics[width=1.0\linewidth]{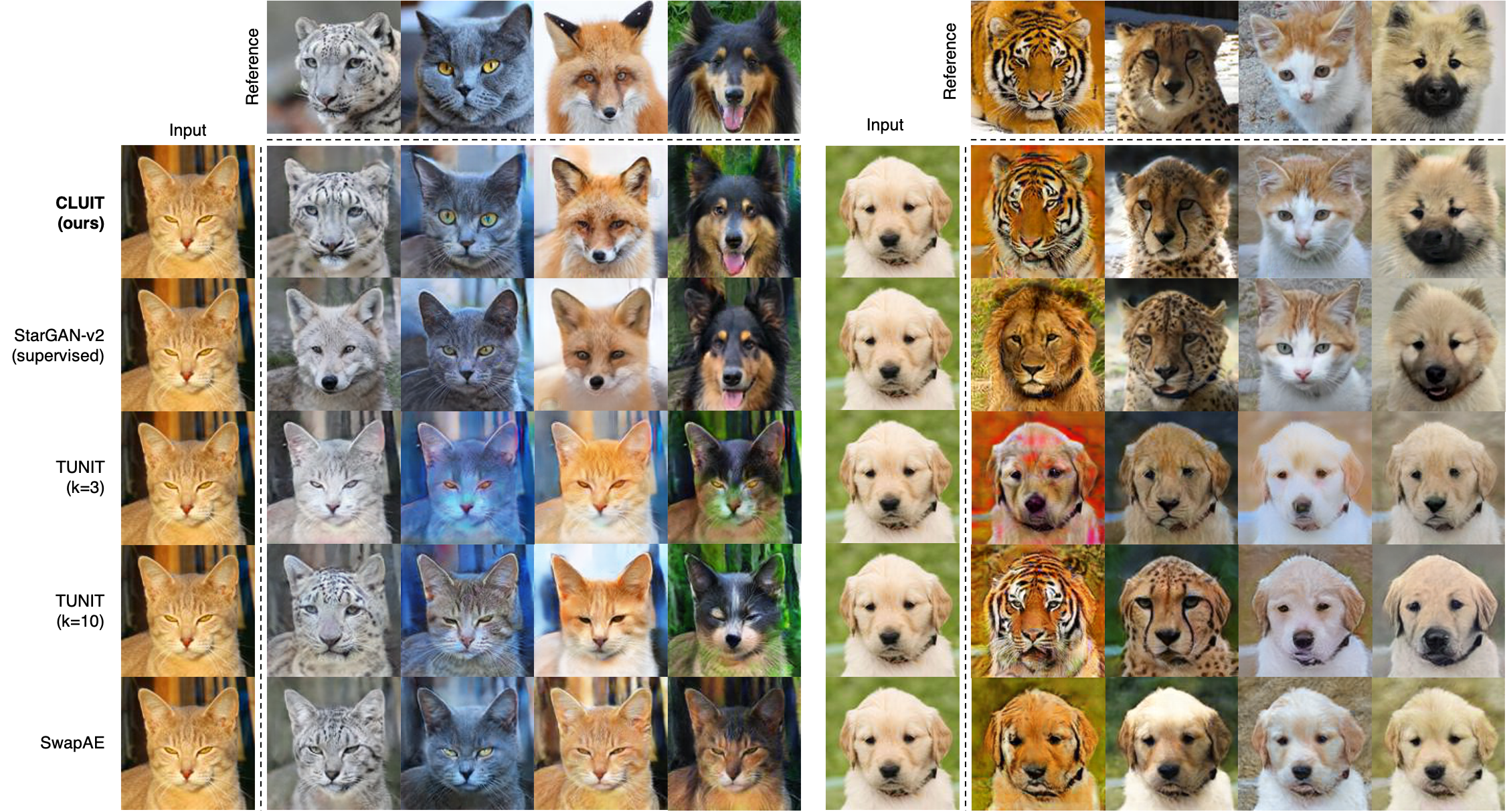}
    \caption{Qualitative comparison of the proposed method and other baselines in AFHQ dataset. Each method translates the input images (left-most column of each grid) into corresponding output, reflecting the styles of the reference images (top-most row). The images are generated by each method specified in the left.}
    \label{fig:comparison-afhq}
\end{figure*}

    \vspace{0.2cm}

    \noindent \textbf{Evaluation protocol.}
    All the methods translate an input image into its corresponding output given a single reference image.
    We use the test set images from each dataset as the input images and each input image is translated using 10 reference images randomly sampled from the test set resulting in 10 output images.
    We use the same set of random reference images for all the competing approaches.

    We evaluate both the visual quality and the translation accuracy of generated images.
    To measure the visual quality of generated images, we adopt Frechet inception distance (FID) \cite{heusel2017gans} which empirically estimates the distribution of real and generated images using pre-trained Inception-v3 \cite{simonyan2014very}.
    In AFHQ dataset, we report the mean of class-wise FID using the ground-truth labels of reference images to penalize the case when a model produces realistic reconstructions of input images.

    Furthermore, we report the translation accuracy measured by the pre-trained ResNet-50 \cite{he2016deep} classifier to determine whether the translation output belongs to the reference image's domain.
    For CelebA dataset, we select 12 binary attributes (see Figure \ref{fig:attribute} for more details) that are considered invariant to the structure of images so that we can verify the transferability of the pose-invariant style.

    \vspace{0.2cm}

    \noindent \textbf{Baselines.}
    We use TUNIT \cite{baek2020rethinking} and SwapAE \cite{park2020swapping} as our unsupervised baselines.
    In the case of TUNIT, the number of clusters must be specified in advance, but it is difficult to know the optimal number for each dataset.
    Therefore, we train the model with varying numbers ($k = 3, 10, 30$) of clusters.
    We also report the results of StarGAN-v2 \cite{choi2020stargan} in AFHQ dataset to confirm the results of the leading supervised method.

    \vspace{0.2cm}

    \noindent \textbf{Implementation.}
    We set $\lambda_{\text{cyc}} = 1$ and $\lambda_{\text{ct}}^D = \lambda_{\text{ct}}^G = 0.1$ in Equation \ref{eq:full}.
    We use $M = 4$ patches in all experiments except for the contrastive style match ablation.
    The batch size is set to 32.
    We adopt the non-saturating adversarial loss \cite{goodfellow2014generative} with lazy $R_1$ regularization \cite{mescheder2018training} using $\gamma = 1$ for CelebA and $\gamma = 10$ for AFHQ.
    We use Adam \cite{kingma2014adam} optimizer with $\beta_1 = 0$ and $\beta_2 = 0.99$.
    The learning rate is set to $5 \times 10^{-5}$.
    We train the model for 100,000 iterations in total.

\begin{figure*}[!htbp]
    \centering
    \includegraphics[width=1.0\linewidth]{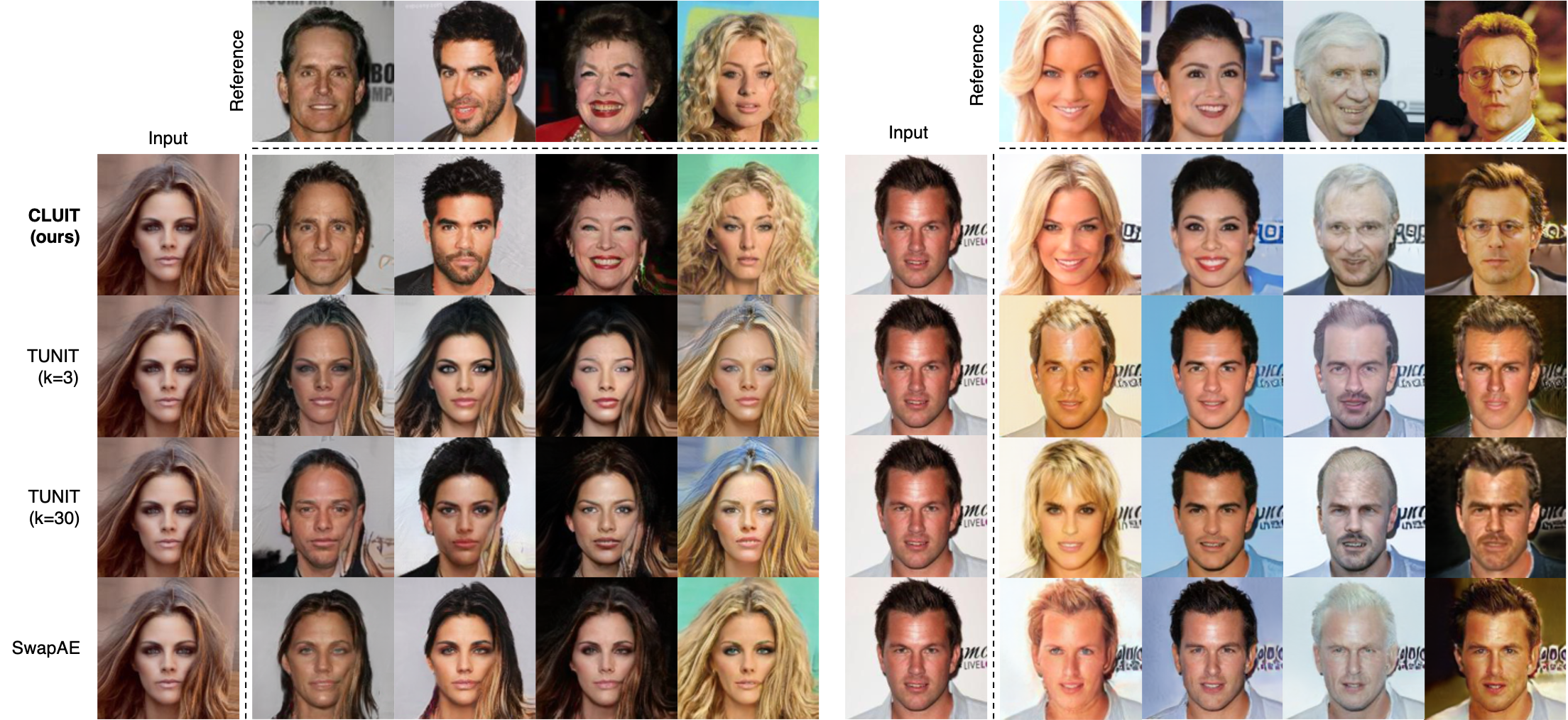}
    \caption{Qualitative comparison of the proposed method and other baselines in CelebA dataset. Each method translates the input images (left-most column of each grid) into corresponding output, reflecting the styles of the reference images (top-most row). The images are generated by each method specified in the left.}
    \label{fig:comparison-celeba}
\end{figure*}

    \subsection{Comparison to baselines}

        The representative results of CLUIT are presented in Figure \ref{fig:teaser}.
        The results show that CLUIT can successfully transform the input images into corresponding outputs, reflecting distinctive visual styles of reference images.

        Figure \ref{fig:comparison-afhq} illustrates the comparison results with the baselines in AFHQ dataset.
        All the methods generate clear images, but they differ significantly in terms of the style being transferred.
        As expected, StarGAN-v2 successfully transfers the style of reference images since it utilizes the ground truth labels to learn class-specific distribution.
        However, we observe that StarGAN-v2 often fails to capture intra-class styles if the images are coarsely labeled.
        For example, the "wildlife" class contains a much wider variety of animals than the other classes (i.e., "cat" and "dog").
        In this case, StarGAN-v2 translates an image into a rather random species in the "wildlife" category, not reflecting a given reference image.
        In Figure \ref{fig:comparison-afhq}, StarGAN-v2 translates a cat into a fox even if a leopard is given as a reference.
        
        The results of TUNIT have high variances depending on the number of pre-defined clusters.
        When the number of clusters is small ($k = 3$), TUNIT only allows the local texture or color change rather than overall appearance.
        Even with the best performing configuration ($k = 10$), the model often fails to differentiate between visually similar species such as cats and foxes.
        In the case of SwapAE, only the color and texture distributions are matched with the reference images.
        On the other hand, our method successfully transfers the distinctive features, including overall shapes and local details (e.g., shape of a face part, fur pattern, skin color, etc.).

        \begin{table}
            \centering
            \begin{tabular}{|l|r|r|r|r|}
                \hline
                    \multirow[b]{2}{*}{Method} & \multicolumn{2}{|c|}{AFHQ} & \multicolumn{2}{c|}{CelebA} \\
                \cline{2-5}
                     & mFID$\downarrow$ & Acc$\uparrow$ & FID$\downarrow$ & Acc$\uparrow$ \\
                \hline
                \hline
                    StarGan-v2* & 14.6 & 97.0 & - & - \\
                \hline
                    TUNIT-3 & 82.1 & 40.5 & 8.3 & 77.3 \\
                    TUNIT-10 & 39.5 & 77.6 & 5.3 & 78.6 \\
                    TUNIT-30 & 48.9 & 66.7 & 6.1 & 81.7 \\
                    SwapAE & 65.3 & 52.5 & 6.2 & 77.5 \\
                    CLUIT (ours) & \textbf{17.8} & \textbf{96.1} & \textbf{5.0} & \textbf{90.1} \\
                \hline
            \end{tabular}
            \caption{Quantitative comparison of the proposed method and other baselines. StarGAN-v2* is a supervised model trained with ground-truth labels. $K$ in TUNIT-$K$ means the predefined number of clusters. $\uparrow$ means larger numbers are better, $\downarrow$ means smaller numbers are better.}
            \label{tab:quantitative}
        \end{table}

        As shown in Table \ref{tab:quantitative}, our method outperforms all the unsupervised baselines by a large margin even comparable to the supervised baseline, in terms of class-specific visual quality (mFID) and translation accuracy.

\begin{figure}[!tbp]
    \centering
    \includegraphics[width=1.0\linewidth]{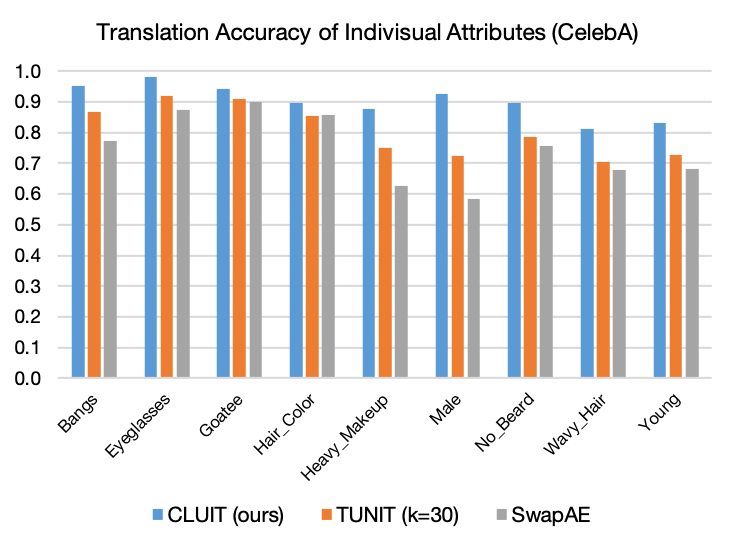}
    \caption{Translation accuracy of individual attributes in CelebA dataset. 12 attributes considered to be invariant of different poses are selected. Note that we show the average accuracy of hair color related attributes for brevity. They are BlackHair, BlondHair, BrownHair, and GrayHair.}
    \label{fig:attribute}
\end{figure}

        For CelebA (Figure \ref{fig:comparison-celeba}), our method renders distinctive visual features of reference images in various poses.
        For other unsupervised baselines, the visual features are conveyed, but they are rather partial or incomplete, often resulting in unrealistic images.
        For example, all the methods modify the hair color following the reference images but only CLUIT succeeds in drawing other hair details such as hair texture and hair length (e.g., long blond wavy hair or short black wavy hair).
        Other details, including eyeglasses, beards, wrinkles, and make-up styles are also successfully transferred to render different views with consistent styles rather than random blends.

        Improvements in terms of both FID and translation accuracy support our observation (Table \ref{tab:quantitative}).
        Figure \ref{fig:attribute} illustrates the translation accuracy of individual attributes.
        We observe that the abstract style-related attributes (e.g., gender or age) show larger margins than relatively low-level attributes (e.g., hair color) or regional attributes (e.g., goatee).
        This observation demonstrates that CLUIT captures and transfers higher-level visual styles via contrastive learning.

    \subsection{Analysis and ablation study}

        In this section, we first analyze the latent style space to verify whether the style encoder learns a smooth and semantically meaningful feature space without any direct constraints.
        We also conduct an ablation study to understand the effect of different data augmentation schemes and patch aggregation methods in contrastive style match.

        \vspace{0.2cm}

        \noindent \textbf{Latent style space analysis.}
        There are two desired properties for latent style codes.
        First, they should learn meaningful visual features.
        Also, the learned visual features should be invariant to different structures. 
        To figure out if these properties are satisfied, we conduct two additional analysis tasks; similarity search and linear interpolation.

        \begin{figure}[!tp]
            \centering
            \includegraphics[width=1.0\linewidth]{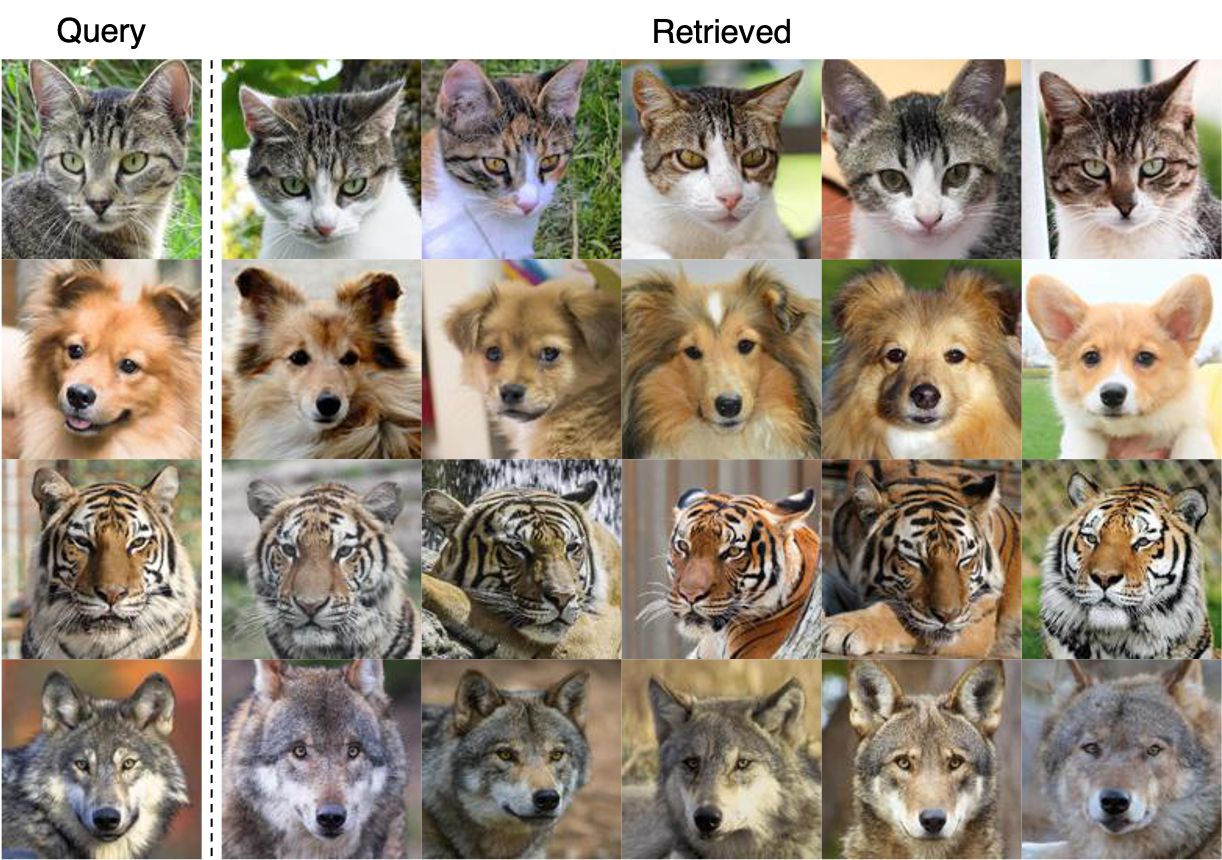}
            \caption{Similarity search results on AFHQ dataset. The first column shows query images, and the next five columns show the search results sorted by the cosine similarity of latent style codes. We can observe that style codes successfully represent high-level class information such as species.}
            \label{fig:retrieval}
        \end{figure}

        For similarity search, all test images are encoded into corresponding latent style codes and normalized to calculate the cosine similarity between pairs of images.
        Figure \ref{fig:retrieval} illustrates the images sorted by the similarity with given query images.
        We observe that even if the pose or background is different, images of the same species and similar style as the query image are retrieved.
        The results illustrate that the style codes capture such visual features invariant to structural changes.

\begin{figure}[!tp]
            \centering
            \includegraphics[width=1.0\linewidth]{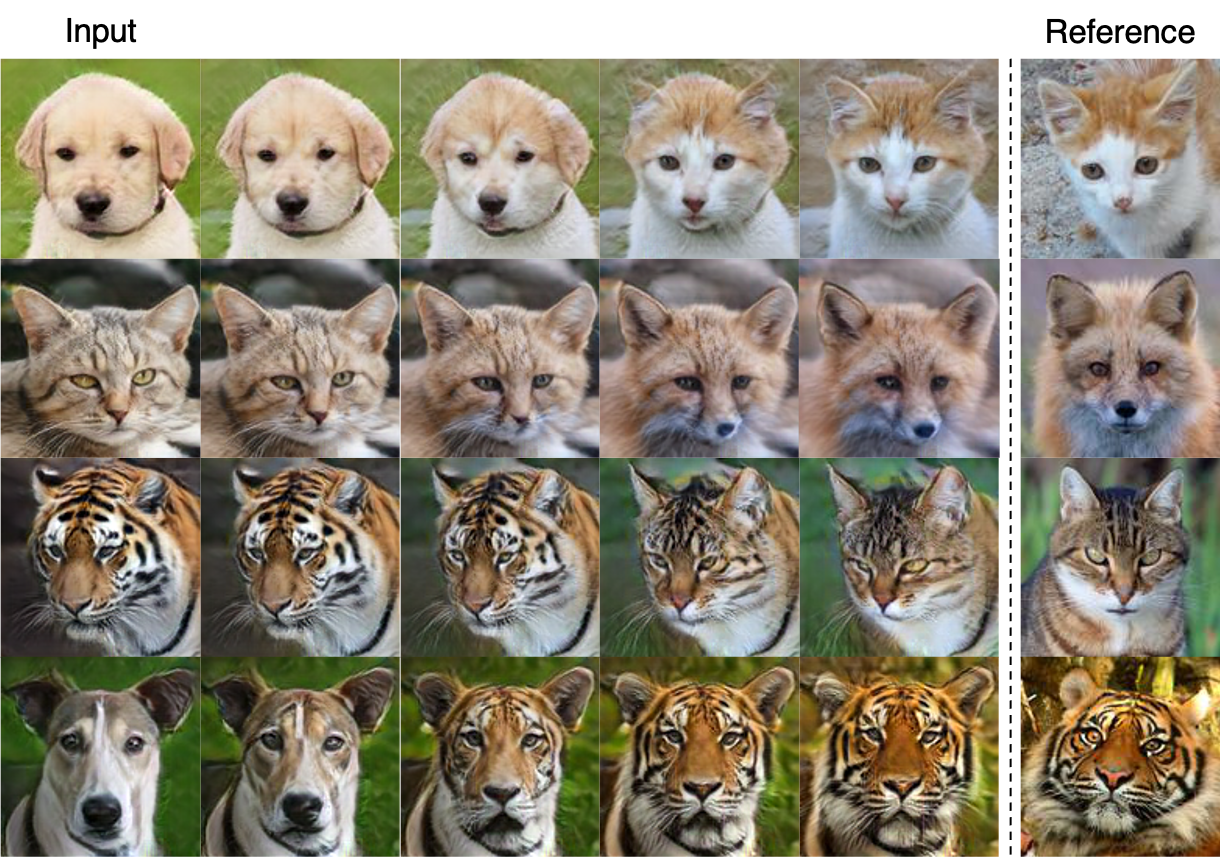}
            \caption{Linear interpolation result on AFHQ dataset. The first and last columns show the input and reference images, respectively. The middle four columns show images generated by interpolating the style codes of the input and the reference while keeping the input image fixed. The results indicate that as the portion of the style code from the reference increases, high-level visual features such as ear and nose shapes gradually change.}
            \label{fig:interpolation}
        \end{figure}

        Figure \ref{fig:interpolation} demonstrates the linear interpolation results by keeping the input image fixed while interpolating the style code between the input and given reference image.
        The interpolated samples smoothly change towards the reference species while preserving the overall structure, such as the position of face parts and the head direction.

        \begin{table}
            \centering
            \begin{tabular}{|l|c|r|r|}
                \hline
                    Augmentation & Crop Size & mFID$\downarrow$ & Accuracy$\uparrow$ \\
                \hline
                \hline
                    Color & 0.125 - 1.0 & 21.25 & 94.9 \\
                    Color + Affine & 0.125 - 1.0 & \textbf{17.84} & 96.1 \\
                    Color + Affine & 0.5 - 1.0 & 18.12 & \textbf{96.6} \\
                    Color + Affine & 0.9 - 1.0 & 20.71 & 95.8 \\
                \hline
            \end{tabular}
            \caption{Performance of CLUIT with different data augmentation schemes. The results indicate that CLUIT benefits from affine transformations and access to small crop size.}
            \label{tab:augmentation}
        \end{table}

        \vspace{0.2cm}

        \noindent \textbf{Data augmentation.}
        Table \ref{tab:augmentation} shows the effect of different data augmentation schemes.
        We start with a set of transformations that includes random cropping, horizontal flipping, and color distortion, following the state-of-the-art unsupervised representation learning \cite{chen2020simple}.
        While affine transformations are known to be relatively less useful with contrastive learning, we observe that our model benefits from them in the context of image-to-image translation.
        This is because the translation task aims to transfer geometry-invariant features to images with an arbitrary pose.
        Thus, geometric transformations help the model to generalize better.
        The affine transformations used for experiments are rotation, shear, and shift transformation.
        Random crop size also affects the results.
        Access to small size patches (i.e., 1/8 of image size) helps the model learn local patterns such as eye shape and fur pattern.
        Using only large patches decreases the visual quality and translation accuracy.
        We use random cropping with a crop size of 0.125-1.0, horizontal flipping, affine transformation, and color transformation as our default set of transformations for all experiments.

        \begin{table}
            \centering
            \begin{tabular}{|l|r|r|}
                \hline
                    Matching Strategy & mFID$\downarrow$ & Accuracy$\uparrow$ \\
                \hline
                \hline
                    Full Image & 19.08 & 95.8 \\
                    PatchMean-1 & 20.43 & 93.0 \\
                    PatchMean-4 & \textbf{17.84} & 96.1 \\
                    PatchMean-8 & 17.98 & \textbf{96.4} \\
                \hline
            \end{tabular}
            \caption{Performance of CLUIT with different contrastive style match schemes. $K$ in PatchMean-$K$ denotes the number of patches sampled for patch aggregation. The results show that the patch aggregation are effective not only for mitigating structure loss problem, but also for translation capabilities.}
            \label{tab:matching}
        \end{table}

        \vspace{0.2cm}

\begin{figure}[!tbp]
    \centering
    \includegraphics[width=1.0\linewidth]{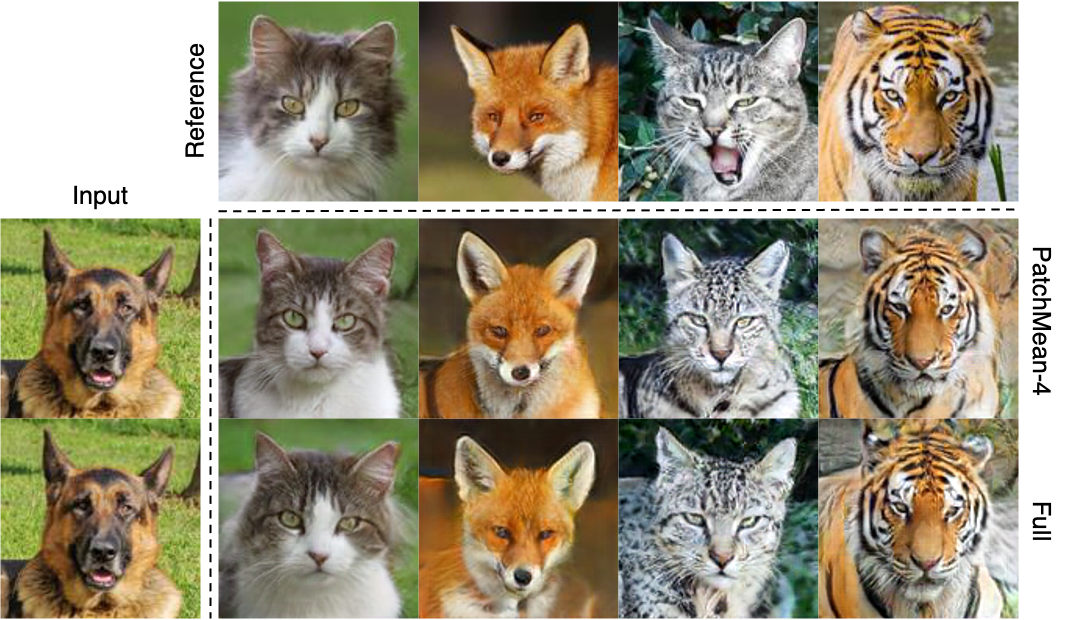}
    \caption{Comparison results of different style-matching strategies. The second line is the result of applying the patch aggregation, and the third line used the representation of the full image match. Without the patch aggregation, the output often tends to be a modified version of the reference, ignoring the structure of the input.}
    \label{fig:matching}
\end{figure}

        \noindent \textbf{Contrastive style match.}
        Figure \ref{fig:matching} illustrates that the patch aggregation effectively mitigates the structure loss in the output images.
        When calculating the style match loss using the representation of the full images, the resulting image is more related to the structure of the reference images (e.g., head size, body position) rather than maintaining the structure of the input images.
        Conversely, when patch aggregation is applied, the output images consistently retain the structure of the input images.
        In addition, loss of structure tends to introduce more artifacts, resulting in slight performance degradation (see Table \ref{tab:matching}).
        We also confirm that using more than a certain amount of patches will benefit performance.





\section{Conclusion}

    In this paper, we introduce an unsupervised image-to-image translation method that learns to transfer distinctive styles between images.
    To this end, we designed a specialized multi-task discriminator that learns meaningful visual representation via contrastive learning.
    The image translation process is supervised by the learned visual representation of the discriminator therefore does not require domain labels or their estimates.
    The proposed framework sets the model free from errors propagated from inconsistent or coarse labels or inaccurate estimation of class labels, leading to better visual quality and translation accuracy.
    Experimental results show that our CLUIT outperforms the state-of-the-art unsupervised image-to-image translation methods and is even comparable to the leading supervised method.

{\small \bibliographystyle{ieee_fullname} \bibliography{egbib}}

\begin{thebibliography}{10}\itemsep=-1pt

\bibitem{bachman2019learning}
Philip Bachman, R~Devon Hjelm, and William Buchwalter.
\newblock Learning representations by maximizing mutual information across
  views.
\newblock In {\em Advances in Neural Information Processing Systems}, pages
  15535--15545, 2019.

\bibitem{baek2020rethinking}
Kyungjune Baek, Yunjey Choi, Youngjung Uh, Jaejun Yoo, and Hyunjung Shim.
\newblock Rethinking the truly unsupervised image-to-image translation.
\newblock {\em arXiv preprint arXiv:2006.06500}, 2020.

\bibitem{bahng2020exploring}
Hyojin Bahng, Sunghyo Chung, Seungjoo Yoo, and Jaegul Choo.
\newblock Exploring unlabeled faces for novel attribute discovery.
\newblock pages 5821--5830, 2020.

\bibitem{chen2020simple}
Ting Chen, Simon Kornblith, Mohammad Norouzi, and Geoffrey Hinton.
\newblock A simple framework for contrastive learning of visual
  representations.
\newblock {\em arXiv preprint arXiv:2002.05709}, 2020.

\bibitem{choi2018stargan}
Yunjey Choi, Minje Choi, Munyoung Kim, Jung-Woo Ha, Sunghun Kim, and Jaegul
  Choo.
\newblock Stargan: Unified generative adversarial networks for multi-domain
  image-to-image translation.
\newblock In {\em Proceedings of the IEEE conference on computer vision and
  pattern recognition}, pages 8789--8797, 2018.

\bibitem{choi2020stargan}
Yunjey Choi, Youngjung Uh, Jaejun Yoo, and Jung-Woo Ha.
\newblock Stargan v2: Diverse image synthesis for multiple domains.
\newblock In {\em Proceedings of the IEEE/CVF Conference on Computer Vision and
  Pattern Recognition}, pages 8188--8197, 2020.

\bibitem{goodfellow2014generative}
Ian Goodfellow, Jean Pouget-Abadie, Mehdi Mirza, Bing Xu, David Warde-Farley,
  Sherjil Ozair, Aaron Courville, and Yoshua Bengio.
\newblock Generative adversarial nets.
\newblock In {\em Advances in neural information processing systems}, pages
  2672--2680, 2014.

\bibitem{hadsell2006dimensionality}
Raia Hadsell, Sumit Chopra, and Yann LeCun.
\newblock Dimensionality reduction by learning an invariant mapping.
\newblock In {\em 2006 IEEE Computer Society Conference on Computer Vision and
  Pattern Recognition (CVPR'06)}, volume~2, pages 1735--1742. IEEE, 2006.

\bibitem{he2020momentum}
Kaiming He, Haoqi Fan, Yuxin Wu, Saining Xie, and Ross Girshick.
\newblock Momentum contrast for unsupervised visual representation learning.
\newblock In {\em Proceedings of the IEEE/CVF Conference on Computer Vision and
  Pattern Recognition}, pages 9729--9738, 2020.

\bibitem{he2015delving}
Kaiming He, Xiangyu Zhang, Shaoqing Ren, and Jian Sun.
\newblock Delving deep into rectifiers: Surpassing human-level performance on
  imagenet classification.
\newblock In {\em Proceedings of the IEEE international conference on computer
  vision}, pages 1026--1034, 2015.

\bibitem{he2016deep}
Kaiming He, Xiangyu Zhang, Shaoqing Ren, and Jian Sun.
\newblock Deep residual learning for image recognition.
\newblock In {\em Proceedings of the IEEE conference on computer vision and
  pattern recognition}, pages 770--778, 2016.

\bibitem{heusel2017gans}
Martin Heusel, Hubert Ramsauer, Thomas Unterthiner, Bernhard Nessler, and Sepp
  Hochreiter.
\newblock Gans trained by a two time-scale update rule converge to a local nash
  equilibrium.
\newblock In {\em Advances in neural information processing systems}, pages
  6626--6637, 2017.

\bibitem{hinton2006reducing}
Geoffrey~E Hinton and Ruslan~R Salakhutdinov.
\newblock Reducing the dimensionality of data with neural networks.
\newblock {\em science}, 313(5786):504--507, 2006.

\bibitem{huang2017arbitrary}
Xun Huang and Serge Belongie.
\newblock Arbitrary style transfer in real-time with adaptive instance
  normalization.
\newblock In {\em Proceedings of the IEEE International Conference on Computer
  Vision}, pages 1501--1510, 2017.

\bibitem{karras2019style}
Tero Karras, Samuli Laine, and Timo Aila.
\newblock A style-based generator architecture for generative adversarial
  networks.
\newblock In {\em Proceedings of the IEEE conference on computer vision and
  pattern recognition}, pages 4401--4410, 2019.

\bibitem{kim2017learning}
Taeksoo Kim, Moonsu Cha, Hyunsoo Kim, Jung~Kwon Lee, and Jiwon Kim.
\newblock Learning to discover cross-domain relations with generative
  adversarial networks.
\newblock In {\em International Conference on Machine Learning}, pages
  1857--1865, 2017.

\bibitem{kingma2014adam}
Diederik~P Kingma and Jimmy Ba.
\newblock Adam: A method for stochastic optimization.
\newblock {\em arXiv preprint arXiv:1412.6980}, 2014.

\bibitem{larsson2017colorization}
Gustav Larsson, Michael Maire, and Gregory Shakhnarovich.
\newblock Colorization as a proxy task for visual understanding.
\newblock In {\em Proceedings of the IEEE Conference on Computer Vision and
  Pattern Recognition}, pages 6874--6883, 2017.

\bibitem{li2018closed}
Yijun Li, Ming-Yu Liu, Xueting Li, Ming-Hsuan Yang, and Jan Kautz.
\newblock A closed-form solution to photorealistic image stylization.
\newblock In {\em Proceedings of the European Conference on Computer Vision
  (ECCV)}, pages 453--468, 2018.

\bibitem{liu2017unsupervised}
Ming-Yu Liu, Thomas Breuel, and Jan Kautz.
\newblock Unsupervised image-to-image translation networks.
\newblock In {\em Advances in neural information processing systems}, pages
  700--708, 2017.

\bibitem{liu2019few}
Ming-Yu Liu, Xun Huang, Arun Mallya, Tero Karras, Timo Aila, Jaakko Lehtinen,
  and Jan Kautz.
\newblock Few-shot unsupervised image-to-image translation.
\newblock In {\em Proceedings of the IEEE International Conference on Computer
  Vision}, pages 10551--10560, 2019.

\bibitem{liu2019exploiting}
Xialei Liu, Joost Van De~Weijer, and Andrew~D Bagdanov.
\newblock Exploiting unlabeled data in cnns by self-supervised learning to
  rank.
\newblock {\em IEEE transactions on pattern analysis and machine intelligence},
  41(8):1862--1878, 2019.

\bibitem{liu2015deep}
Ziwei Liu, Ping Luo, Xiaogang Wang, and Xiaoou Tang.
\newblock Deep learning face attributes in the wild.
\newblock In {\em Proceedings of the IEEE international conference on computer
  vision}, pages 3730--3738, 2015.

\bibitem{luan2017deep}
Fujun Luan, Sylvain Paris, Eli Shechtman, and Kavita Bala.
\newblock Deep photo style transfer.
\newblock In {\em Proceedings of the IEEE Conference on Computer Vision and
  Pattern Recognition}, pages 4990--4998, 2017.

\bibitem{maas2013rectifier}
Andrew~L Maas, Awni~Y Hannun, and Andrew~Y Ng.
\newblock Rectifier nonlinearities improve neural network acoustic models.
\newblock In {\em Proc. icml}, volume~30, page~3, 2013.

\bibitem{mescheder2018training}
Lars Mescheder, Andreas Geiger, and Sebastian Nowozin.
\newblock Which training methods for gans do actually converge?
\newblock {\em arXiv preprint arXiv:1801.04406}, 2018.

\bibitem{mirza2014conditional}
Mehdi Mirza and Simon Osindero.
\newblock Conditional generative adversarial nets.
\newblock {\em arXiv preprint arXiv:1411.1784}, 2014.

\bibitem{noroozi2016unsupervised}
Mehdi Noroozi and Paolo Favaro.
\newblock Unsupervised learning of visual representations by solving jigsaw
  puzzles.
\newblock In {\em European Conference on Computer Vision}, pages 69--84.
  Springer, 2016.

\bibitem{noroozi2017representation}
Mehdi Noroozi, Hamed Pirsiavash, and Paolo Favaro.
\newblock Representation learning by learning to count.
\newblock In {\em Proceedings of the IEEE International Conference on Computer
  Vision}, pages 5898--5906, 2017.

\bibitem{oord2018representation}
Aaron van~den Oord, Yazhe Li, and Oriol Vinyals.
\newblock Representation learning with contrastive predictive coding.
\newblock {\em arXiv preprint arXiv:1807.03748}, 2018.

\bibitem{park2020swapping}
Taesung Park, Jun-Yan Zhu, Oliver Wang, Jingwan Lu, Eli Shechtman, Alexei
  Efros, and Richard Zhang.
\newblock Swapping autoencoder for deep image manipulation.
\newblock {\em Advances in Neural Information Processing Systems}, 33, 2020.

\bibitem{pathak2016context}
Deepak Pathak, Philipp Krahenbuhl, Jeff Donahue, Trevor Darrell, and Alexei~A
  Efros.
\newblock Context encoders: Feature learning by inpainting.
\newblock In {\em Proceedings of the IEEE conference on computer vision and
  pattern recognition}, pages 2536--2544, 2016.

\bibitem{santa2017deeppermnet}
Rodrigo Santa~Cruz, Basura Fernando, Anoop Cherian, and Stephen Gould.
\newblock Deeppermnet: Visual permutation learning.
\newblock In {\em Proceedings of the IEEE Conference on Computer Vision and
  Pattern Recognition}, pages 3949--3957, 2017.

\bibitem{simonyan2014very}
Karen Simonyan and Andrew Zisserman.
\newblock Very deep convolutional networks for large-scale image recognition.
\newblock {\em arXiv preprint arXiv:1409.1556}, 2014.

\bibitem{ulyanov2016instance}
Dmitry Ulyanov, Andrea Vedaldi, and Victor Lempitsky.
\newblock Instance normalization: The missing ingredient for fast stylization.
\newblock {\em arXiv preprint arXiv:1607.08022}, 2016.

\bibitem{vincent2008extracting}
Pascal Vincent, Hugo Larochelle, Yoshua Bengio, and Pierre-Antoine Manzagol.
\newblock Extracting and composing robust features with denoising autoencoders.
\newblock In {\em Proceedings of the 25th international conference on Machine
  learning}, pages 1096--1103, 2008.

\bibitem{yi2017dualgan}
Zili Yi, Hao Zhang, Ping Tan, and Minglun Gong.
\newblock Dualgan: Unsupervised dual learning for image-to-image translation.
\newblock In {\em Proceedings of the IEEE international conference on computer
  vision}, pages 2849--2857, 2017.

\bibitem{yoo2019photorealistic}
Jaejun Yoo, Youngjung Uh, Sanghyuk Chun, Byeongkyu Kang, and Jung-Woo Ha.
\newblock Photorealistic style transfer via wavelet transforms.
\newblock In {\em Proceedings of the IEEE International Conference on Computer
  Vision}, pages 9036--9045, 2019.

\bibitem{zhu2017unpaired}
Jun-Yan Zhu, Taesung Park, Phillip Isola, and Alexei~A Efros.
\newblock Unpaired image-to-image translation using cycle-consistent
  adversarial networks.
\newblock In {\em Proceedings of the IEEE international conference on computer
  vision}, pages 2223--2232, 2017.

\end{thebibliography}


\cleardoublepage
\appendix
\noindent{\Large\bfseries Appendix}
\vspace{0.5cm}

In this supplementary material, we provide implementation details as well as additional experimental results. Implementation details include detailed network architecture and training details for better reproducibility. For additional experimental results, we provide additional samples of the proposed method, comparison results with baselines, and failure cases.


\section{Implementation Details}

    In this section, we present the architectural details of CLUIT consisting of three modules, a discriminator, a generator, and a style encoder.
    All the network weights are initialized using He initialization \cite{he2015delving}. 

    \subsection{Architecture}

        \noindent \textbf{Discriminator.}
        Our discriminator consists of 5 shared pre-activation residual blocks, followed by two separate output branches, each producing the contrastive representation and the adversarial logit.
        Each output branch contains 2 fully connected layers where the dimension of contrastive representation is set to 256 and the dimension for real/fake classification is set to 1.
        We use leaky ReLU \cite{maas2013rectifier} activation for all layers but did not use other normalization techniques.

        \vspace{0.2cm}

        \noindent \textbf{Generator.}
        Basically, we use Adaptive Instance Normalization (AdaIN) \cite{huang2017arbitrary} based locality-preserving architecture \cite{choi2020stargan, karras2019style} as our generator.
        Our generator consists of 3 downsample layers, 4 intermediate layers, and 3 upsample layers.
        We use Instance Normalization (IN) \cite{ulyanov2016instance} for the downsample layers and AdaIn for the upsample layers.
        For intermediate layers, the first half of layers are normalized with IN, and the remaining layers are normalized with AdaIN.
        The latent style code is injected into all AdaIN layers to stylize the output. We use leaky ReLU activation for all layers.

        \vspace{0.2cm}

        \noindent \textbf{Style Encoder.}
        The style encoder consists of 5 pre-activation residual blocks followed by two fully connected layers.
        The output dimension of style encoder is set to 128.
        We use leaky ReLU activation for all layers without normalization layers.

        \begin{table}
            \centering
            \begin{tabular}{|c|c|c|c|}
                \hline
                    Type & Layer & Resample & Output Shape \\
                \hline
                    - & Image X & - & 128x128x3 \\
                    - & Conv1x1 & - & 128x128x64 \\
                    shared & ResBlk & AvgPool & 64x64x128 \\
                    shared & ResBlk & AvgPool & 32x32x256 \\
                    shared & ResBlk & AvgPool & 16x16x512 \\
                    shared & ResBlk & AvgPool & 8x8x512 \\
                    shared & ResBlk & AvgPool & 4x4x512 \\
                    unshared & LReLU & - & 4x4x512 \\
                    unshared & Conv4x4 & - & 1x1x512 \\
                    unshared & LReLU & - & 1x1x512 \\
                    unshared & Reshape & - & 512 \\
                    unshared & Linear & - & 256, 1 \\
                \hline
            \end{tabular}
            \caption{Discriminator architecture.}
            \label{tab:appendix-arch-d}
        \end{table}

        \begin{table}
            \centering
            \begin{tabular}{|c|c|c|c|c|}
                \hline
                    Layer & Resample & Norm & Output Shape \\
                \hline
                    Image X & - & - & 128x128x3 \\
                    Conv1x1 & - & - & 128x128x64 \\
                    ResBlk & AvgPool & IN & 64x64x128 \\
                    ResBlk & AvgPool & IN & 32x32x256 \\
                    ResBlk & AvgPool & IN & 16x16x512 \\
                    ResBlk & - & IN & 16x16x512 \\
                    ResBlk & - & IN & 16x16x512 \\
                    ResBlk & - & AdaIN & 16x16x512 \\
                    ResBlk & - & AdaIN & 16x16x512 \\
                    ResBlk & Upsample & AdaIn & 32x32x256 \\
                    ResBlk & Upsample & AdaIn & 64x64x128 \\
                    ResBlk & Upsample & AdaIn & 128x128x64 \\
                    Conv1x1 & - & - & 128x128x3 \\
                \hline
            \end{tabular}
            \caption{Generator architecture.}
            \label{tab:appendix-arch-g}
        \end{table}

        \begin{table}
            \centering
            \begin{tabular}{|c|c|c|}
                \hline
                    Layer & Resample & Output Shape \\
                \hline
                    Image X & - & 128x128x3 \\
                    Conv1x1 & - & 128x128x64 \\
                    ResBlk & AvgPool & 64x64x128 \\
                    ResBlk & AvgPool & 32x32x256 \\
                    ResBlk & AvgPool & 16x16x512 \\
                    ResBlk & AvgPool & 8x8x512 \\
                    ResBlk & AvgPool & 4x4x512 \\
                    LReLU & - & 4x4x512 \\
                    Conv4x4 & - & 1x1x512 \\
                    LReLU & - & 1x1x512 \\
                    Reshape & - & 512 \\
                    Linear & - & 128 \\
                \hline
            \end{tabular}
            \caption{Style encoder architecture.}
            \label{tab:appendix-arch-e}
        \end{table}

    \subsection{Training details}

        Here we provide the full details needed to train the model. We set $\lambda_{cyc}=1$, $\lambda_{ct}^D=\lambda_{ct}^G=0.1$ as weights for each loss term. We use $M=4$ patches in all experiments except for the contrastive style match ablation. When computing $\mathcal{L}_{\text{ct}}^D$, we apply a random cropping transform to the input image with the same size range used for the augmentation to match the granularity.
        The temperature $\tau$ is set to 0.07 \cite{he2020momentum} and the size of the negative dictionary is set to 2048.
        The batch size is set to 32. We adopt the non-saturating adversarial loss \cite{goodfellow2014generative} with lazy $R_1$ regularization \cite{mescheder2018training} using $\gamma=1$ for CelebA and $\gamma=10$ for AFHQ. We use Adam \cite{kingma2014adam} optimizer with $\beta_1=0$ and $\beta_2=0.99$.
        The learning rate is set to $5 \times 10^{-5}$.
        For the evaluation, we adopt exponential moving averages over parameters of all modules.
        We use Resnet-50 as an oracle classifier while the classifier is trained using the training images of each dataset. The classifier accuracy is 99.7 in AFHQ and 93.2 in CelebA (average of 12 binary attributes).


\section{Additional Results}

    In this section, we show additional results on both AFHQ and CelebA dataset.

    \subsection{Additional results of CLUIT}

\begin{figure*}[p]
    \centering
    \captionsetup[subfigure]{labelformat=empty}
    \begin{subfigure}{\textwidth}
        \centering
        \includegraphics[width=0.85\linewidth]{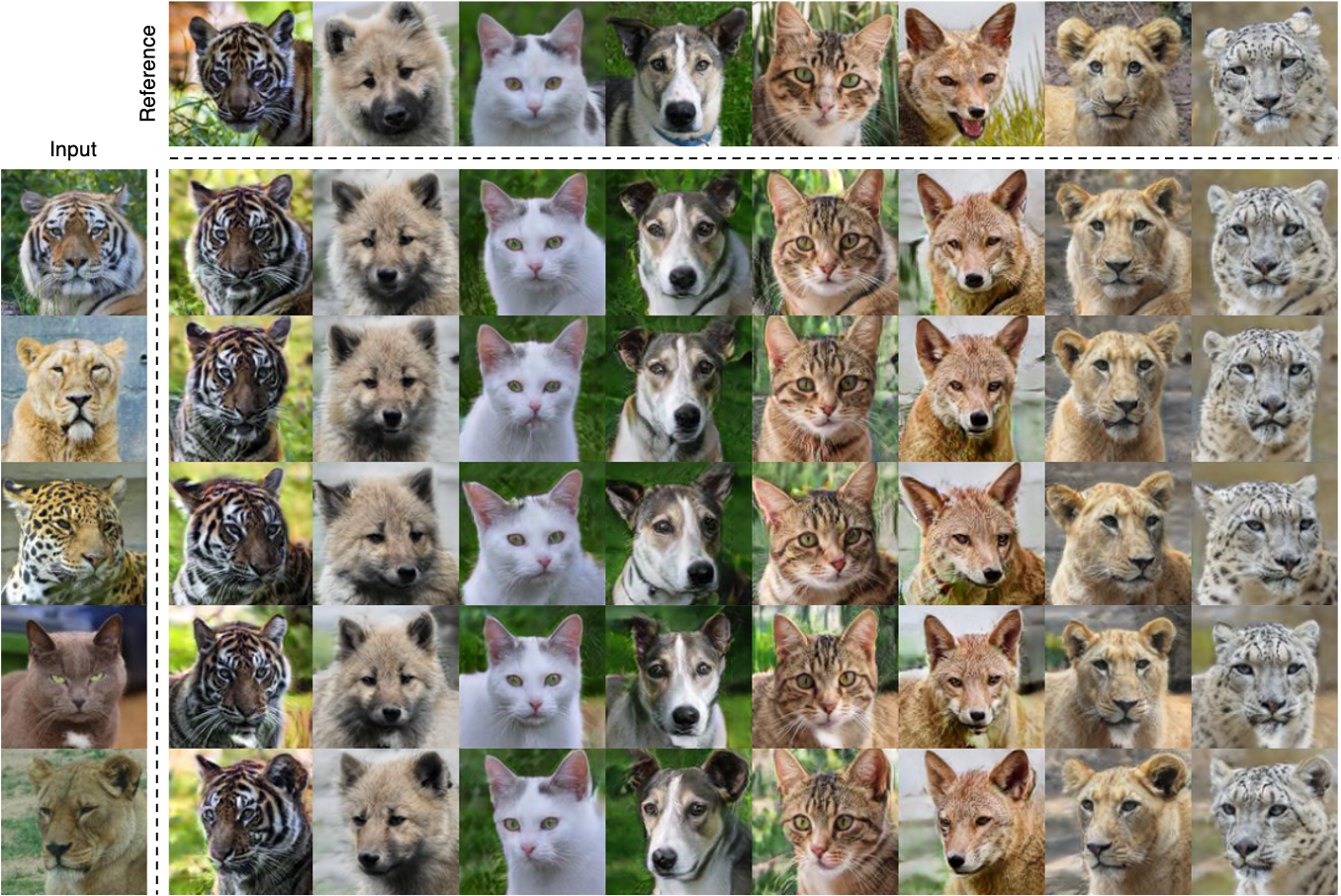}
        \caption{}
    \end{subfigure}
    \begin{subfigure}{\textwidth}
        \centering
        \includegraphics[width=0.85\linewidth]{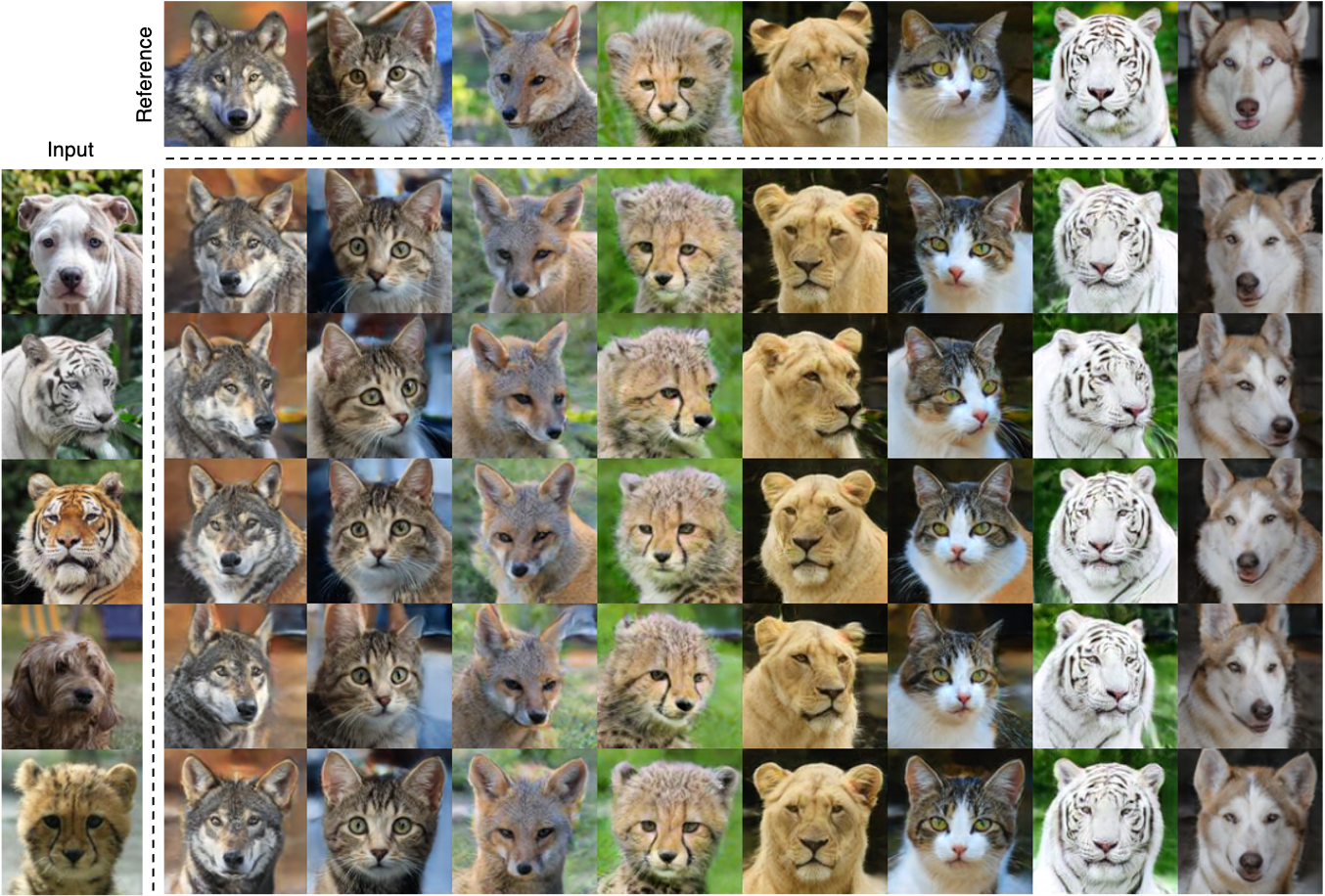}
        \caption{}
    \end{subfigure}
    \vspace{-0.4cm}
    \caption{Image-to-image translation results on AFHQ. The first column shows the input images and the first row of each grid shows the reference images, while the rest of the images are generated by the proposed method, CLUIT. CLUIT captures and transfers high-level semantics such as fur color and pattern, and face shapes.}
    \label{fig:appendix-cluit-afhq}
\end{figure*}

\begin{figure*}[p]
    \centering
    \captionsetup[subfigure]{labelformat=empty}
    \begin{subfigure}{\textwidth}
        \centering
        \includegraphics[width=0.85\linewidth]{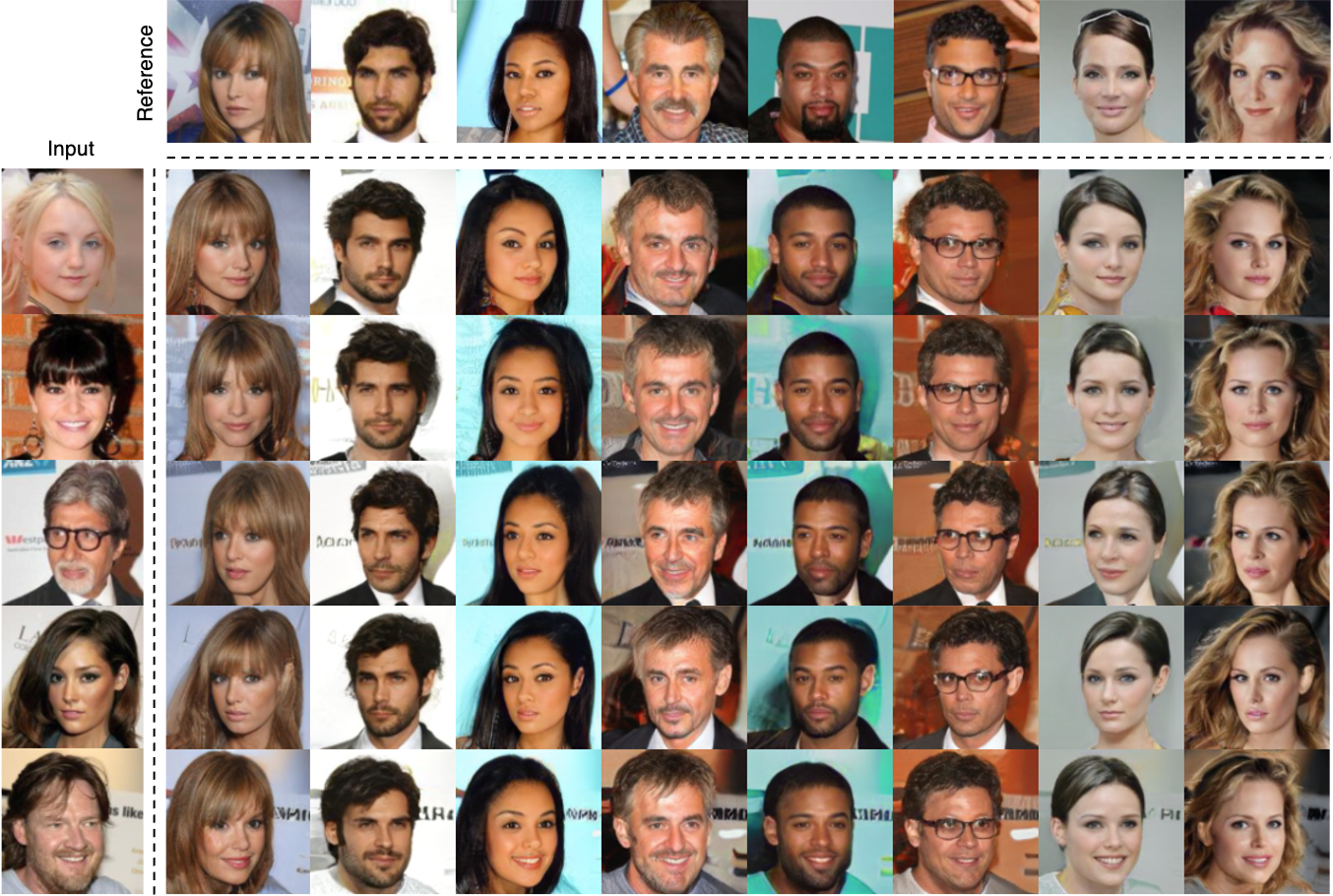}
        \caption{}
    \end{subfigure}
    \begin{subfigure}{\textwidth}
        \centering
        \includegraphics[width=0.85\linewidth]{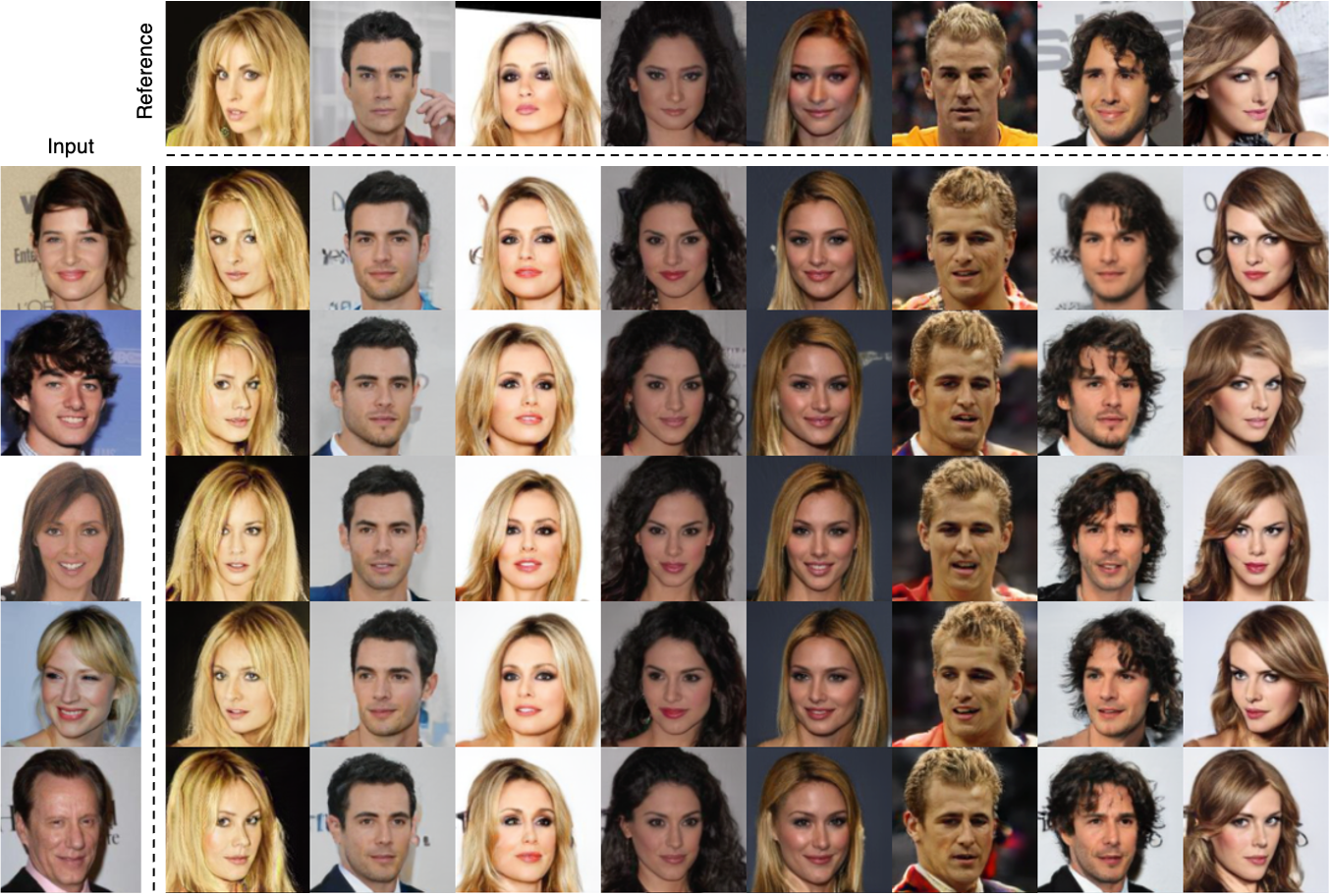}
        \caption{}
    \end{subfigure}
    \vspace{-0.4cm}
    \caption{Image-to-image translation results on CelebA. The first column shows the input images and the first row of each grid shows the reference images, while the rest of the images are generated by the proposed method, CLUIT. CLUIT captures and transfers high-level semantics such as hair type and color, eyeglasses, beard, and makeup styles.}
    \label{fig:appendix-cluit-celeba}
\end{figure*}

        Figure \ref{fig:appendix-cluit-afhq} and \ref{fig:appendix-cluit-celeba} illustrate additional samples generated by the proposed method, CLUIT. In AFHQ, CLUIT successfully synthesizes images by reflecting the unique style of the reference images, including fur pattern, skin color, and facial features. For CelebA, CLUIT renders a distinctive style of the reference image with various poses. Rendered styles include makeup style, glasses, hair texture and color, and more, which form a unique identity.

    \subsection{Additional comparison to existing methods}

\begin{figure*}[p]
    \centering
    \captionsetup[subfigure]{labelformat=empty}
    \begin{subfigure}{\textwidth}
        \centering
        \includegraphics[width=0.75\linewidth]{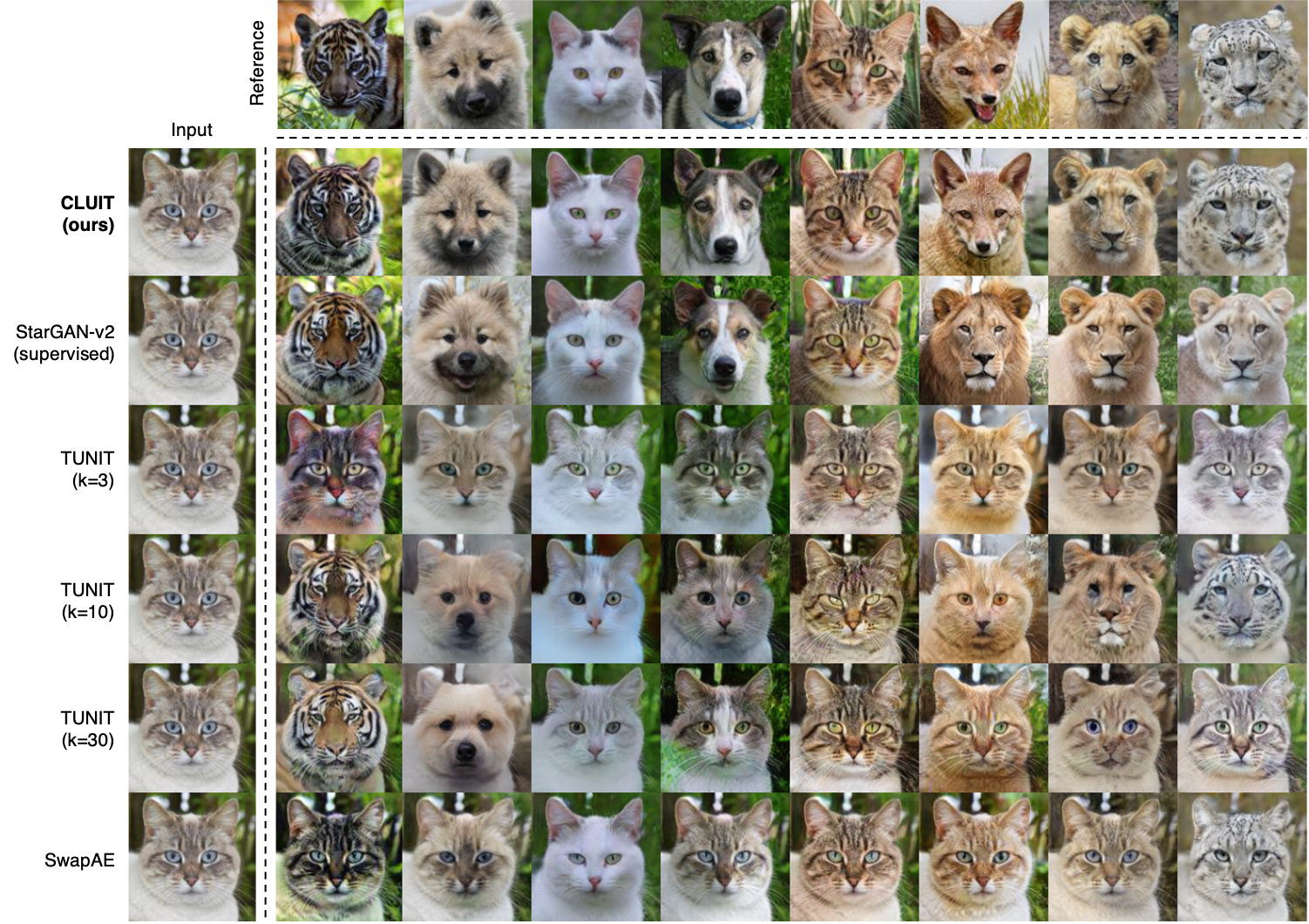}
        \caption{}
    \end{subfigure}
    \begin{subfigure}{\textwidth}
        \centering
        \includegraphics[width=0.75\linewidth]{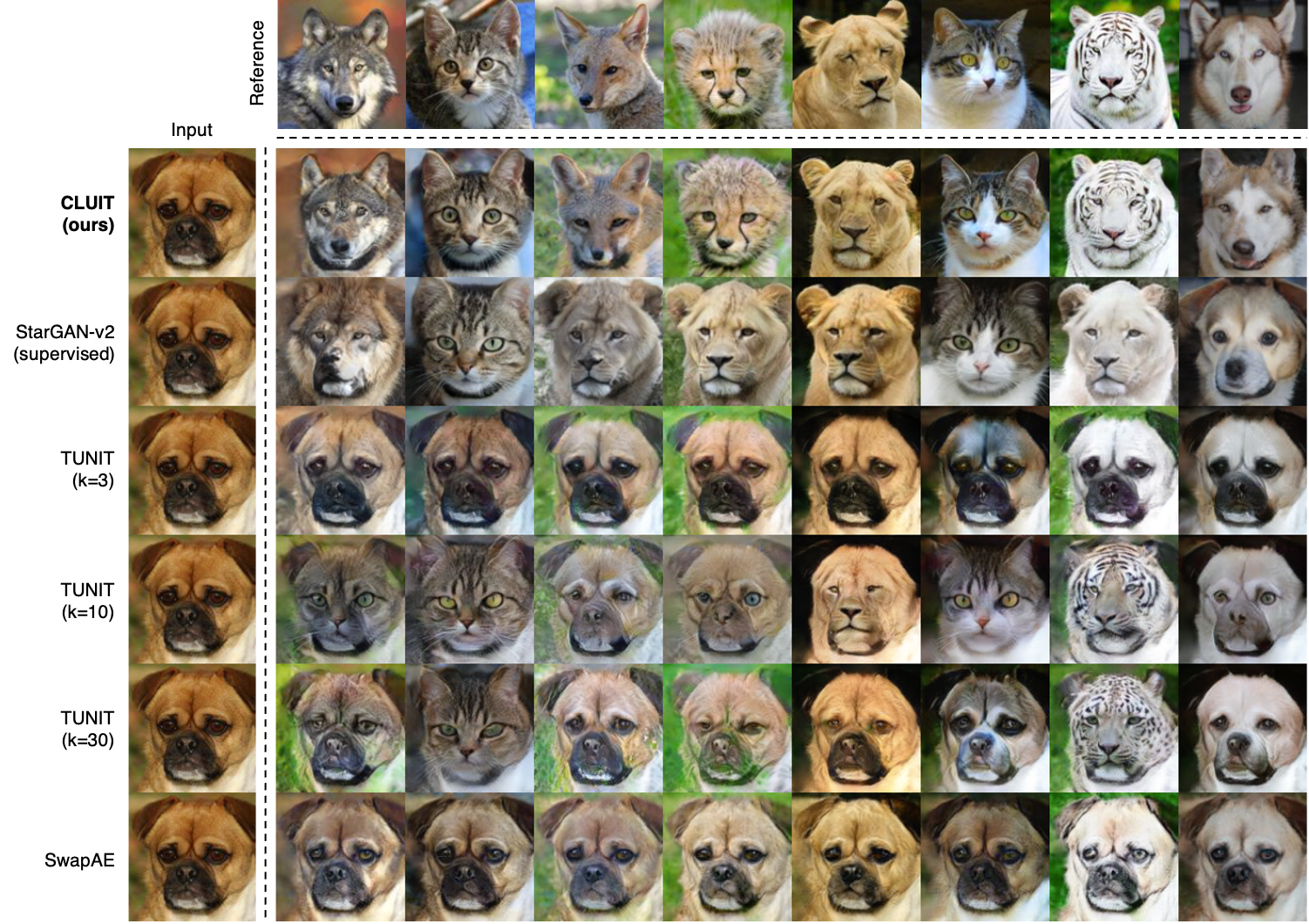}
        \caption{}
    \end{subfigure}
    \vspace{-0.4cm}
    \caption{Qualitative comparison on AFHQ. The first column shows the input images and the first row of each grid shows the reference images, while the rest of the images are generated by each method specified on the left. Note that StarGAN-v2 is a supervised method trained with ground-truth labels, while the others are unsupervised methods. While the baseline methods often generate different species than the reference images, CLUIT successfully renders the distinctive visual styles while preserving the pose of the input image.}
    \label{fig:appendix-comparison-afhq}
\end{figure*}

\begin{figure*}[p]
    \centering
    \captionsetup[subfigure]{labelformat=empty}
    \begin{subfigure}{\textwidth}
        \centering
        \includegraphics[width=0.9\linewidth]{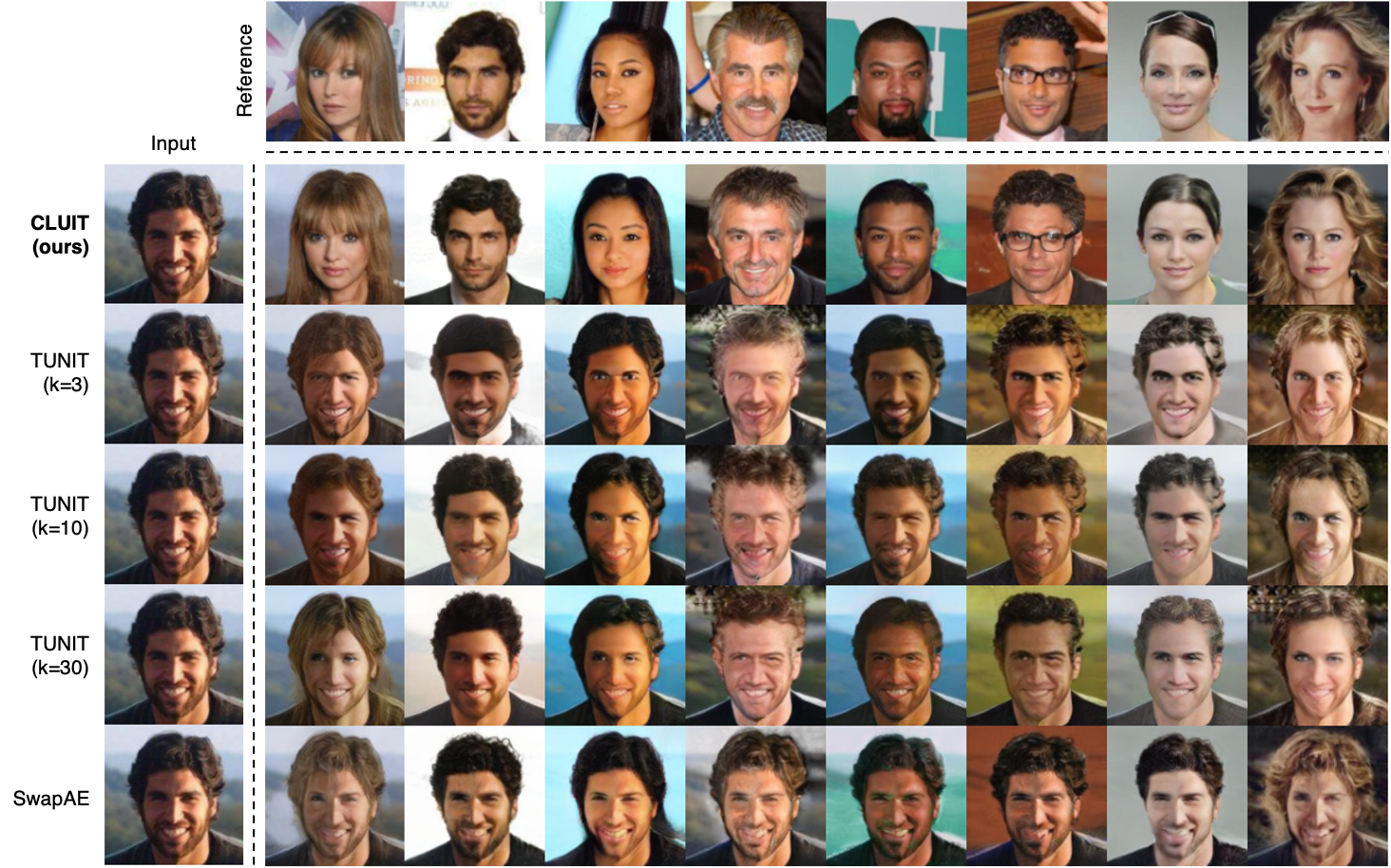}
        \caption{}
    \end{subfigure}
    \begin{subfigure}{\textwidth}
        \centering
        \includegraphics[width=0.9\linewidth]{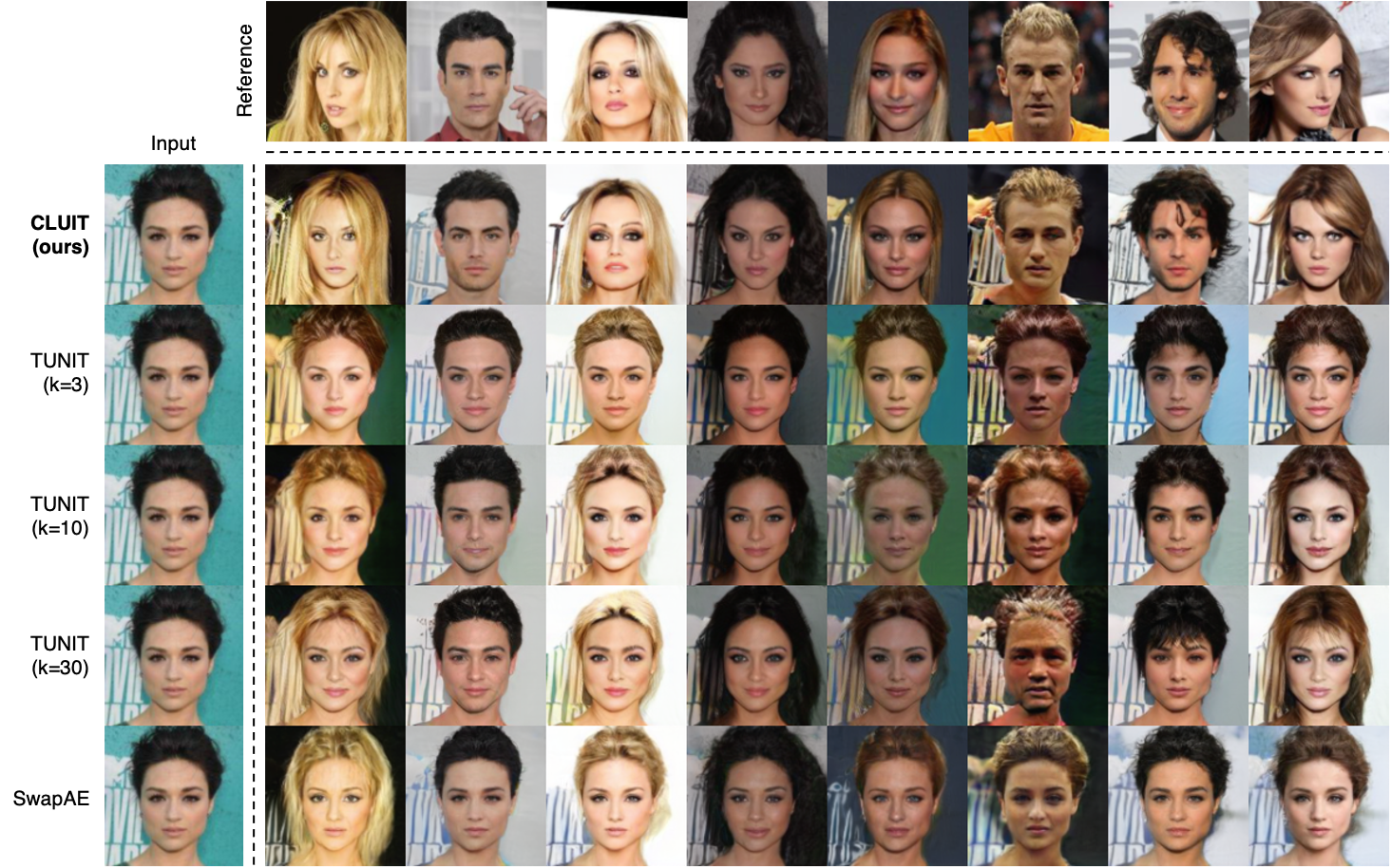}
        \caption{}
    \end{subfigure}
    \vspace{-0.4cm}
    \caption{Qualitative comparison on CelebA. The first column shows the input images and the first row of each grid shows the reference images, while the rest of the images are generated by each method specified on the left. While the baseline methods often produce unrealistic images by partially reflecting the styles of reference images, CLUIT successfully renders the distinctive visual styles of reference images while preserving the pose of the input image.}
    \label{fig:appendix-comparison-celeba}
\end{figure*}

        Figure \ref{fig:appendix-comparison-afhq} and \ref{fig:appendix-comparison-celeba} show additional comparison results with other baselines. The results demonstrate that CLUIT mainly focuses on distinguishing the pose of the subject from the pose-invariant style in the image, while the other baselines try to preserve the contours of the input image, resulting in either incomplete style transfer or local color change.

        We note that the desirable translation results can be set differently according to the user's intention. For example, some may want to change the color or texture, while others may want to reflect the overall look of the reference images. An ideal unsupervised method involves more flexible ways to reflect user's different intents without retraining the model and we leave this as our future work.

    \subsection{Failure Cases}

\begin{figure*}[pht!]
    \centering
    \begin{subfigure}{0.48\textwidth}
        \includegraphics[width=\linewidth]{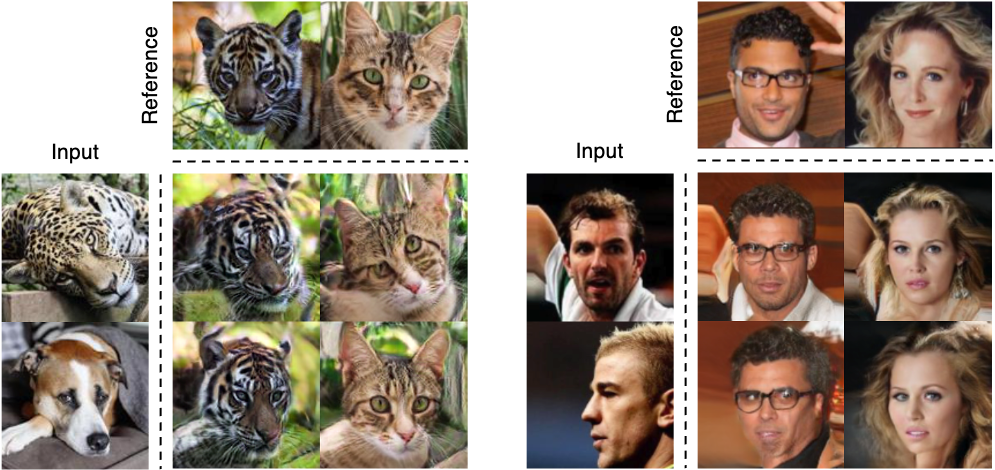}
        \caption{}
    \end{subfigure}
    \hfill%
    \begin{subfigure}{0.48\textwidth}
        \includegraphics[width=\linewidth]{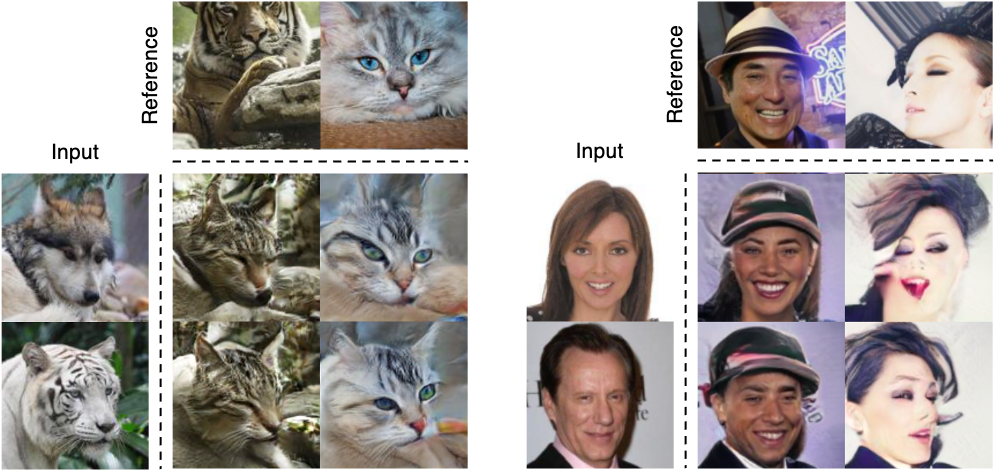}
        \caption{}
    \end{subfigure}
    \caption{(a) Failure cases of CLUIT when the input image of a rare pose are given. (b) Failure cases of CLUIT when the style information in the reference image is limited or partial.}
    \label{fig:appendix-failure}
\end{figure*}

        We also provide several failure cases of the proposed method for further understanding. Figure \ref{fig:appendix-failure} illustrates cases where the model degenerates with rare poses in the input images (Figure \ref{fig:appendix-failure} (a)) and the cases where the reference image's style information is limited or partial (Figure \ref{fig:appendix-failure} (b)).


\end{document}